\documentclass{article}

\PassOptionsToPackage{numbers}{natbib}
\usepackage[preprint]{neurips_2026}

\usepackage[utf8]{inputenc}
\usepackage[T1]{fontenc}
\usepackage{hyperref}
\usepackage{url}
\usepackage{booktabs}
\usepackage{amsfonts}
\usepackage{amsmath}
\usepackage{amssymb}
\usepackage{nicefrac}
\usepackage{microtype}
\usepackage[table]{xcolor}
\usepackage{graphicx}
\usepackage{subcaption}
\usepackage{wrapfig}
\usepackage{multirow}
\usepackage{caption}  %
\usepackage{tikz}     %
\usetikzlibrary{shadows, arrows.meta, positioning, calc}
\definecolor{medblue}{RGB}{34, 60, 86}      %
\definecolor{medbluedark}{RGB}{14, 34, 58}  %

\usepackage{array}
\usepackage{adjustbox}

\newcommand{\figimg}[2][]{%
  \IfFileExists{#2.jpg}{%
    \includegraphics[#1]{#2.jpg}%
  }{%
    \IfFileExists{#2.png}{%
      \includegraphics[#1]{#2.png}%
    }{%
      \IfFileExists{#2}{%
        \includegraphics[#1]{#2}%
      }{%
        \adjustbox{width=\linewidth,height=0.75\linewidth,frame,bgcolor=black!8}{%
          \parbox{\linewidth}{\centering\scriptsize\sffamily\color{black!40}%
            \vspace{0.3\linewidth}%
            \texttt{#2}%
          }%
        }%
      }%
    }%
  }%
}

\newlength{\gridcolwidth}
\newcommand{\gridcell}[2][\gridcolwidth]{%
  \begin{minipage}{#1}\centering
    \figimg[width=\linewidth]{#2}%
  \end{minipage}%
}

\newcommand{\comprow}[1]{%
  \gridcell{figures/comparison/#1-exo} &
  \gridcell{figures/comparison/#1-vista4d} &
  \gridcell{figures/comparison/#1-egox} &
  \gridcell{figures/comparison/#1-egox-diff} &
  \gridcell{figures/comparison/#1-ours} &
  \gridcell{figures/comparison/#1-ours-diff} &
  \gridcell{figures/comparison/#1-gt}%
}

\newcommand{\compheaders}{%
  {\scriptsize Input}
  & {\scriptsize Vista4D}
  & {\scriptsize EgoX}
  & {\scriptsize EgoX error}
  & {\scriptsize \textbf{Ours}}
  & {\scriptsize \textbf{Ours error}}
  & {\scriptsize GT} \\[1.5pt]%
}

\newcommand{\comparisonscenario}[2]{%
  \setlength{\gridcolwidth}{0.135\linewidth}%
  \centering
  \setlength{\tabcolsep}{0.5pt}%
  \resizebox{\linewidth}{!}{%
  \begin{tabular}{c@{\hspace{0.5pt}}c@{\hspace{0.5pt}}c@{\hspace{0.5pt}}c@{\hspace{0.5pt}}c@{\hspace{0.5pt}}c@{\hspace{0.5pt}}c}
    \compheaders
    #2
  \end{tabular}%
  }%
}

\newcommand{\comparisonfigureseenA}{%
  \comparisonscenario{New Activities}{%
    \comprow{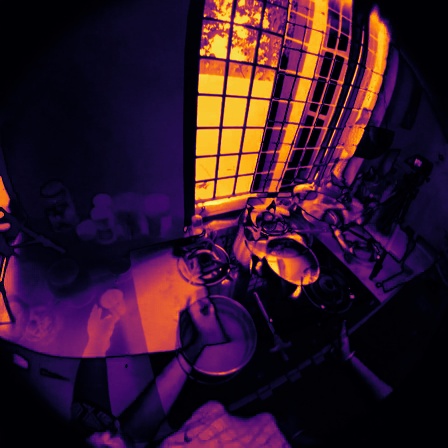} \\[1.5pt]
    \comprow{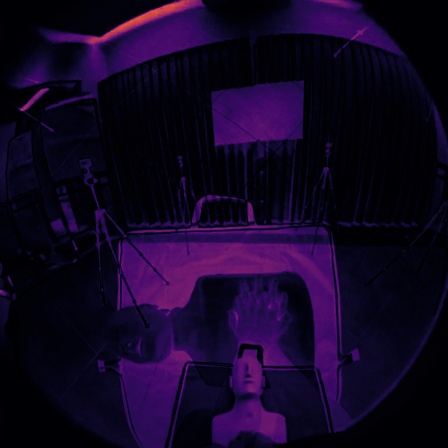} \\[1.5pt]
    \comprow{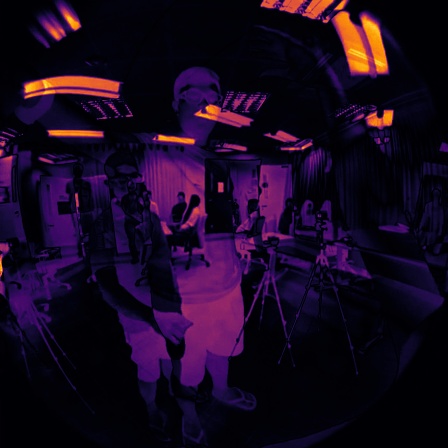}%
  }%
}

\newcommand{\comparisonfigureseensupplemental}{%
  \comparisonscenario{New Activities (supplementary)}{%
    \comprow{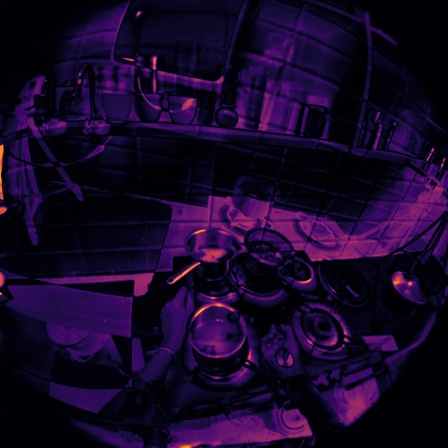} \\[1.5pt]
    \comprow{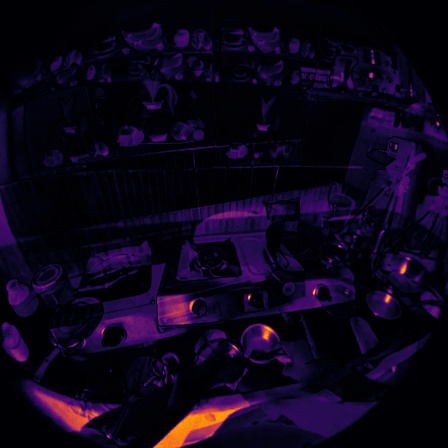} \\[1.5pt]
    \comprow{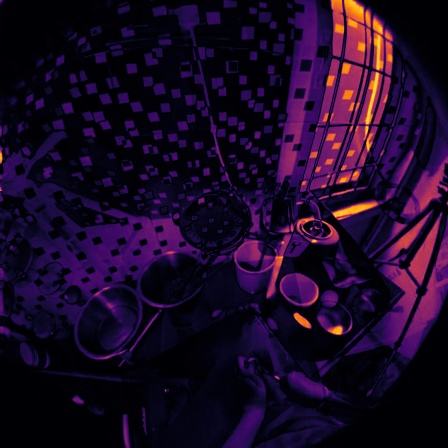} \\[1.5pt]
    \comprow{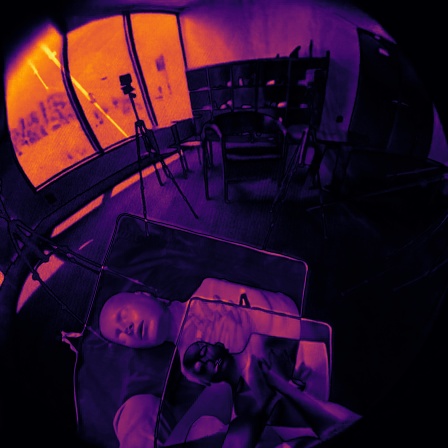} \\[1.5pt]
    \comprow{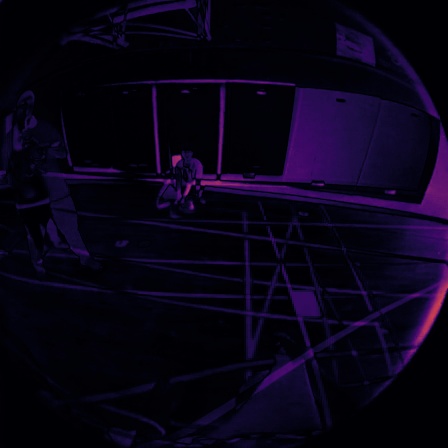}%
  }%
}

\newcommand{\comparisonfigureunseen}{%
  \centering
  \noindent
  \begin{minipage}[c]{16pt}\centering
    \rotatebox{90}{\scriptsize\textbf{EgoExo4D (Unseen)}}%
  \end{minipage}%
  \hspace{2pt}%
  \begin{minipage}[c]{\dimexpr\linewidth-18pt\relax}
    \comparisonscenario{Unseen}{%
      \comprow{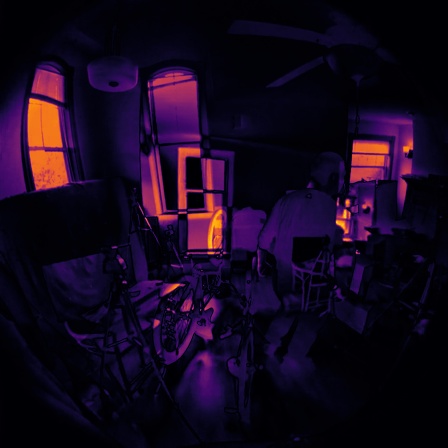} \\[1.5pt]
      \comprow{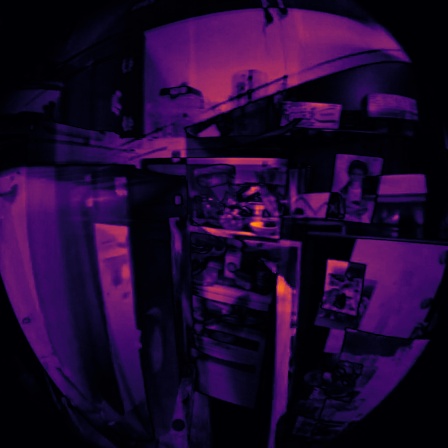}%
    }%
  \end{minipage}%
  \par\vspace{10pt}%
  \noindent
  \begin{minipage}[c]{16pt}\centering
    \rotatebox{90}{\scriptsize\textbf{In-The-Wild}}%
  \end{minipage}%
  \hspace{2pt}%
  \begin{minipage}[c]{\dimexpr\linewidth-18pt\relax}\centering
    \setlength{\tabcolsep}{0.5pt}%
    \begin{tabular}{c@{\hspace{0.5pt}}c@{\hspace{0.5pt}}c}
      {\scriptsize Exo Input}
        & {\scriptsize Ego Rendering}
        & {\scriptsize \textbf{Ours}} \\[1.5pt]
      \includegraphics[width=0.466\linewidth]{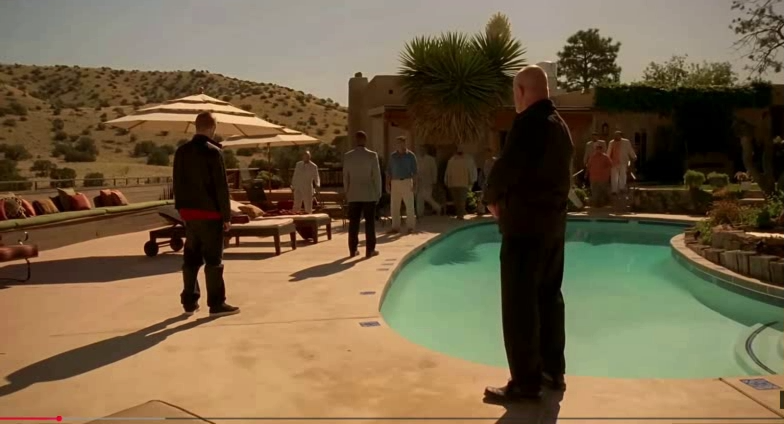} &
      \includegraphics[width=0.266\linewidth]{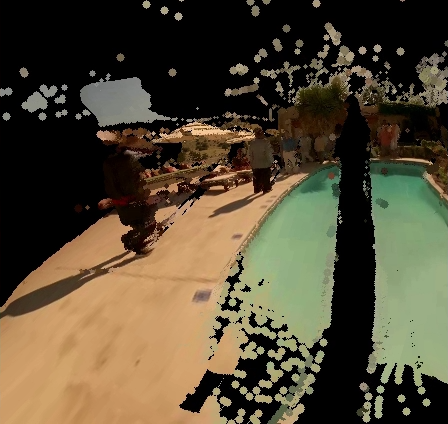} &
      \includegraphics[width=0.266\linewidth]{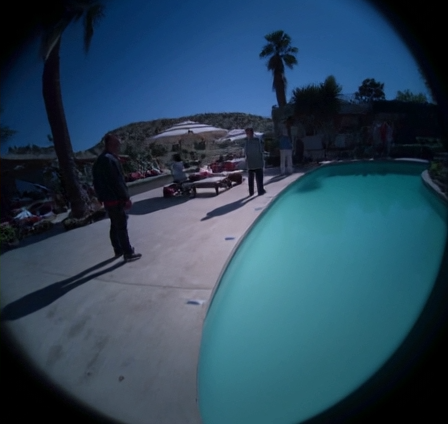} \\[1pt]
      \includegraphics[width=0.466\linewidth]{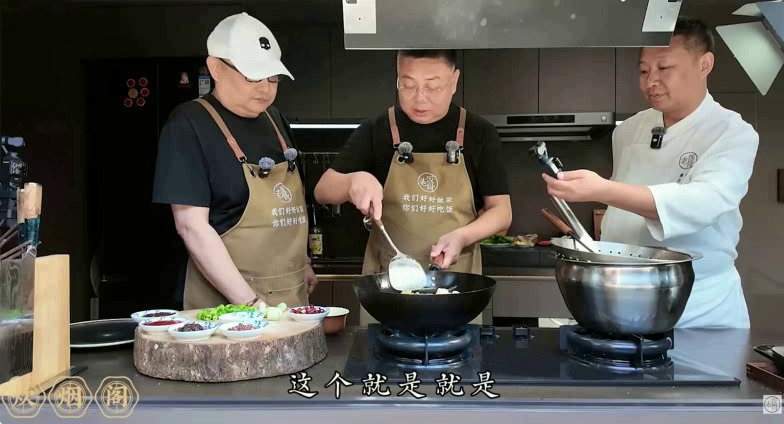} &
      \includegraphics[width=0.266\linewidth]{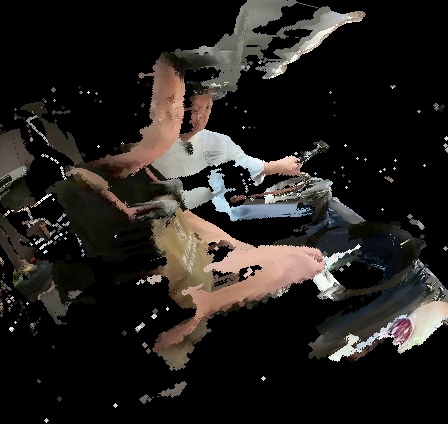} &
      \includegraphics[width=0.266\linewidth]{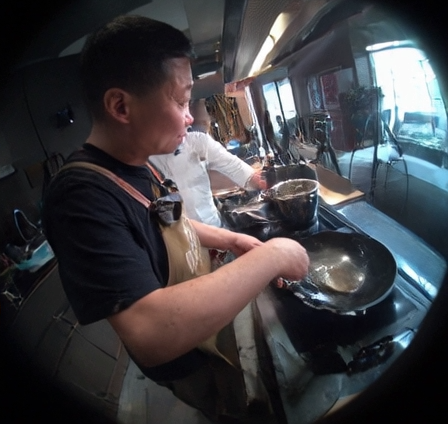}%
    \end{tabular}%
  \end{minipage}%
}

\newcommand{\difficultyplaceholder}[1][]{%
  \begin{tikzpicture}[baseline=0pt]
    \draw[fill=black!12, draw=black!35, line width=0.6pt]
      (0.3pt, 0.3pt)
      rectangle (\dimexpr\linewidth-0.3pt\relax,
                 \dimexpr\linewidth-0.3pt\relax);
  \end{tikzpicture}\\[1pt]%
  {\scriptsize\ifx\\#1\\\strut\else#1\fi}%
}

\newcommand{\difficultyimg}[2]{%
  \begin{tikzpicture}[inner sep=0pt, baseline=(im.south)]
    \node[inner sep=0] (im)
      {\figimg[width=\linewidth]{#1}};
  \end{tikzpicture}\\[1pt]%
  {\scriptsize #2}%
}

\newlength{\difficultypanelside}

\newcommand{\difficultydivider}{%
  \hspace{9pt}%
  \makebox[0pt][c]{\raisebox{5pt}{%
    \begin{tikzpicture}[baseline=(c.center)]
      \node[inner sep=0pt, outer sep=0pt,
            minimum height=\difficultypanelside,
            minimum width=0pt] (c) {};
      \foreach \k in {1,2,3,4,5,6,7,8,9} {%
        \fill[black!75]
          ($(c.south)!\k/10!(c.north)$) circle (0.55pt);
      }
    \end{tikzpicture}%
  }}%
  \hspace{9pt}%
}

\newcommand{\difficultyfig}[1]{%
  \centering
  \setlength{\difficultypanelside}{\dimexpr(\linewidth - 60pt)/6\relax}%
  \begin{minipage}[c]{\difficultypanelside}\centering
    \difficultyimg{figures/comparison/#1-exo}{(a) Exo input}%
  \end{minipage}\difficultydivider%
  \begin{minipage}[c]{\difficultypanelside}\centering
    \difficultyimg{figures/comparison/#1-mesh}{(b) 3D at ego}%
  \end{minipage}\hspace{1pt}%
  \begin{minipage}[c]{\difficultypanelside}\centering
    \difficultyimg{figures/comparison/#1-vista4d}{(c) Vista4D at ego}%
  \end{minipage}\difficultydivider%
  \begin{minipage}[c]{\difficultypanelside}\centering
    \difficultyimg{figures/comparison/#1-egox}{(d) EgoX}%
  \end{minipage}\hspace{1pt}%
  \begin{minipage}[c]{\difficultypanelside}\centering
    \difficultyimg{figures/comparison/#1-ours}{(e) Ours}%
  \end{minipage}\difficultydivider%
  \begin{minipage}[c]{\difficultypanelside}\centering
    \difficultyimg{figures/comparison/#1-gt}{(f) GT}%
  \end{minipage}%
}

\newcommand{\varrow}[1]{%
  \gridcell{figures/comparison_ablation/#1-egox} &
  \gridcell{figures/comparison_ablation/#1-nosem} &
  \gridcell{figures/comparison_ablation/#1-nomask} &
  \gridcell{figures/comparison_ablation/#1-ours} &
  \gridcell{figures/comparison_ablation/#1-gt}%
}

\newcommand{\varheaders}{%
  {\scriptsize EgoX}
  & {\scriptsize\begin{tabular}[b]{@{}c@{}}Ours (no obj tok,\\mask, synth)\end{tabular}}
  & {\scriptsize Ours (no obj mask, synth)}
  & {\scriptsize \textbf{Ours}}
  & {\scriptsize GT} \\[1.5pt]%
}

\newcommand{\varscenario}[1]{%
  \setlength{\gridcolwidth}{0.19\linewidth}%
  \centering
  \setlength{\tabcolsep}{0.5pt}%
  \resizebox{\linewidth}{!}{%
  \begin{tabular}{c@{\hspace{0.5pt}}c@{\hspace{0.5pt}}c@{\hspace{0.5pt}}c@{\hspace{0.5pt}}c}
    \varheaders
    #1
  \end{tabular}%
  }%
}

\newcommand{\comparisonfigureablation}{%
  \varscenario{%
    \varrow{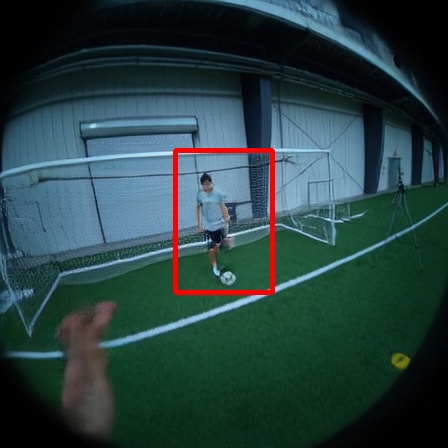} \\[1.5pt]
    \varrow{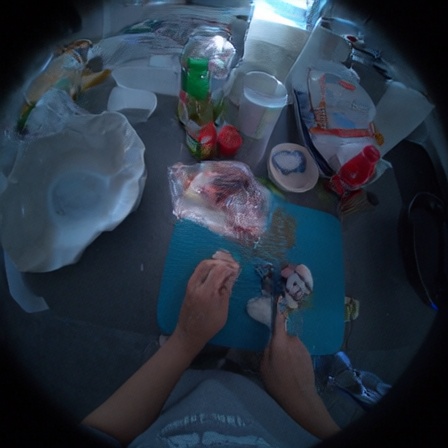}%
  }%
}

\title{Grounded-Exo2Ego: Structured Semantic Grounding for Robust Exocentric-to-Egocentric Video Generation}

\author{%
  {\bfseries Shengze Wang \quad Michael Stengel \quad Tianye Li \quad Seonwook Park} \\
  {\bfseries Amrita Mazumdar \quad Koki Nagano \quad Alex Trevithick \quad Shalini De Mello} \\[3pt]
  {\normalfont NVIDIA} \\[2pt]
  {\normalfont\small\texttt{\{shengzew, mstengel, tianyel, seonwookp,}} \\
  {\normalfont\small\texttt{amritam, knagano, atrevithick, shalinig\}@nvidia.com}} \\[3pt]
  {\normalfont\small\url{https://research.nvidia.com/labs/amri/projects/grounded-exo2ego/}}
}

\begin{document}

\maketitle

\vspace{-0.12in}%
\begin{center}
\includegraphics[width=1.0 \linewidth]{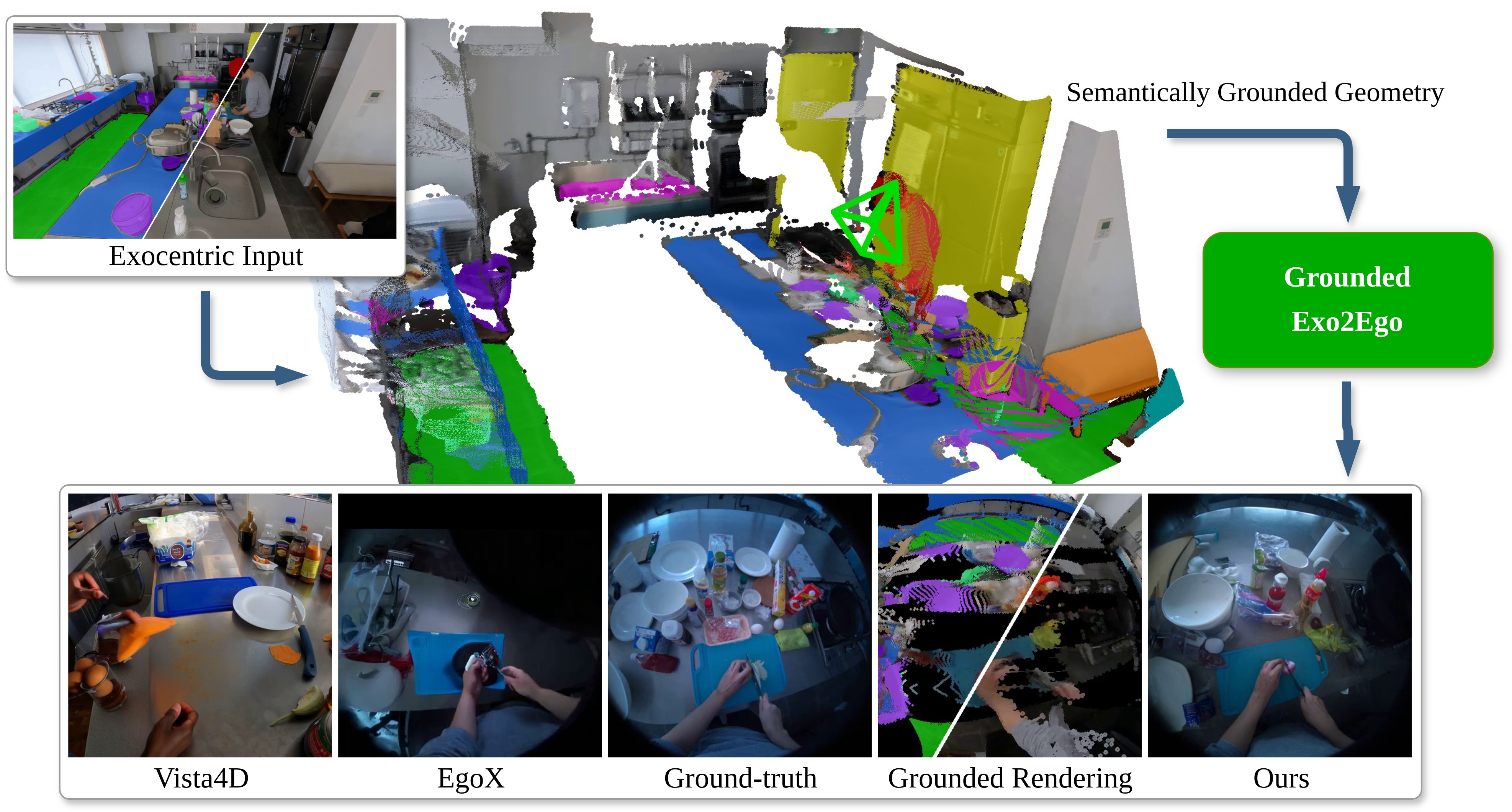}

\captionof{figure}{\textbf{Grounded-Exo2Ego.} From a single exocentric
video (top left), we lift the scene into a semantically grounded 3D reconstruction that anchors a dual-branch video diffusion model.
The generated egocentric views (bottom) are substantially more accurate than those from existing methods.
}
\label{fig:teaser}
\end{center}
\vspace{-6pt}

\begin{abstract}
Generating egocentric video from a single exocentric video is an emerging and
important topic for AR/VR and physical AI. Compared with conventional novel
view synthesis, exo-to-ego generation is a significantly harder task because the standard geometric conditioning becomes highly unreliable under extreme view changes and large unobservable regions.
We present \textbf{Grounded-Exo2Ego}, a principled framework that addresses these
challenges at both the architectural and data levels.
Architecturally, Grounded-Exo2Ego is a dual-branch video diffusion model that
couples a \emph{geometric anchoring} branch, which conditions the generation on the rendering of a 3D reconstruction, with a novel \emph{semantic grounding}
branch, which goes beyond the prevailing geometry-based approach and improves quality by synthesizing challenging regions based on object-level context.
Additionally, we found that the overlooked issue of camera-reconstruction misalignment severely undermines exo-to-ego learning. We thus introduce a camera re-localization algorithm that
resolves this issue and substantially improves quality across all metrics. We further develop
a fully automated synthetic data engine that generates and renders rigged 3D characters in procedurally generated environments.
Evaluation on the challenging EgoExo4D dataset shows that our method outperforms recent state-of-the-art approaches by large margins across all metrics. 
Detailed ablations validate improvements from each of our contributions at both the data and architectural level. 
\end{abstract}

\section{Introduction}
\label{sec:intro}

\begin{figure}[t]
  \centering
  \difficultyfig{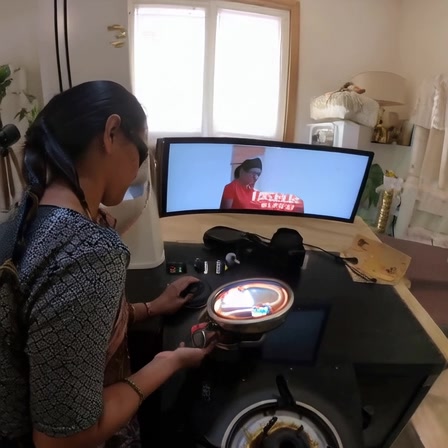}
  \caption{
  \textbf{Exo-to-Ego is inherently more difficult than general view
  synthesis.} State-of-the-art methods, \textit{e.g.} Vista4D~\citep{lin2026vista4d},
  often render monocular 3D reconstruction of the input view (a) from
  a novel view to produce the conditioning video.
  However, such renderings are riddled with holes and corrupted
  geometry at the ego view (b), leading to bad results (c).
  Recent exo-to-ego methods~\citep{egox2025}
  achieve better results (d) but fail to recover the exact
  viewpoint and content.
  }
  \label{fig:difficulty}
\end{figure}

Imagine if one could turn any third-person video into
a first-person view. The vast archive of internet videos
would become a source of egocentric training data at unparalleled
scale and diversity, providing the first-person supervision that
robots and embodied agents urgently need. Important moments from
our own lives, currently frozen in third-person recordings,
could be re-lived in VR. Every tutorial, surgical demonstration,
or sports match could be experienced through the eyes of the performer.
This is what exo-to-ego view synthesis aims to achieve.
Yet despite rapid progress in general view synthesis,
exo-to-ego view synthesis remains practically out of reach for current methods.
Closing this gap requires treating exo-to-ego as a problem in its own right and
solving it systematically across both data and architecture, rather than
transferring machinery built for general view synthesis.

Exo-to-ego is fundamentally harder than conventional novel view
synthesis (NVS) for two coupled reasons.
\textbf{(i)~Geometric conditioning is unreliable.}
Modern view synthesis methods are often conditioned on renderings
of the scene's 3D reconstructions, obtained through monocular
depth estimation on the input video. In exo-to-ego, however,
these signals break down for two compounding reasons:
monocular depth inaccuracies produce stretched surfaces that
appear at the wrong positions in the ego view
(Figure~\ref{fig:difficulty}, b), and large portions of the ego
view are simply not observed from the exo viewpoint. These two issues leave the ego rendering riddled with distortions and large
holes (Figure~\ref{fig:difficulty}, b, c). Even purpose-built
exo-to-ego models fail (e.g.\ EgoX~\citep{egox2025}; Figure~\ref{fig:difficulty}, d) when relying solely on geometric conditioning. These challenging regions need to be
synthesized consistently with the surrounding scene based on semantic context.
\textbf{(ii)~Training data is fragile.} Monocular reconstructions are not aligned with the real-world geometry because of non-rigid distortions and scale/shift ambiguities. As a result, groundtruth cameras calibrated to real-world geometry are also misaligned to monocular reconstructions. Therefore, when existing methods use groundtruth ego camera to render monocular reconstructions during training, much of the rendering (geometric conditioning) is significantly misaligned to the groundtruth during training. This severe misalignment also necessitates heavy filtering, drastically reducing the available training data. For example, EgoX~\citep{egox2025} discards most training samples while the ones left are still misaligned to the groundtruth.

We introduce \textbf{Grounded-Exo2Ego} (Figure~\ref{fig:teaser}), a
principled framework designed to tackle these challenges at both the
\emph{architectural} and the \emph{data} levels. Architecturally,
Grounded-Exo2Ego is a \textbf{dual-branch} video diffusion model
(Figure~\ref{fig:pipeline}) built around two complementary signals.
The \emph{geometric anchoring} branch helps the model determine
the general structure via ego-view renderings of the scene reconstructions.
Going beyond the prevailing geometric-conditioned approach, we introduce the \emph{semantic grounding} branch that further improves object-level generation by anchoring object-level semantics to the scene geometry.
Ego-view renderings of the reconstruction are often so
heavily distorted or occluded that objects become unrecognizable. 
Object-level semantics addresses this problem by providing the
model information about what each region
should depict, helping it restore the correct appearance and location of each object.
At the \emph{data} level, we introduce camera re-localization that improves learning by reducing misalignment between rendering and groundtruth video, and we further develop a fully automated
synthetic data engine that provides accurately annotated training
data. 
As noted above,
monocular reconstructions of exo-view videos are incompatible with the ground-truth ego camera calibrated in the
real world. 
To improve rendering-groundtruth alignment and avoid wasteful filtering, we re-localize the ego
camera inside the inaccurate reconstruction so that the renderings
and ground-truth videos align well for learning.
Beyond real data, our synthetic data generation engine sidesteps the reconstruction quality and alignment issues entirely. Its core is a fully automated data engine 
that produces animatable 3D humans performing realistic actions in randomly generated environments. Besides providing clean training signals, our data
pipeline also fills in the gap for the lack of human-human egocentric 
interaction data by rendering interactions from all participants' perspectives.

\paragraph{To summarize our work:}
\begin{itemize}
  \item We propose \textbf{Grounded-Exo2Ego}, a \emph{dual-branch}
    video diffusion model that couples a \emph{geometric anchoring}
    branch with a novel \emph{semantic grounding} branch, which
    improves generation quality by grounding object-level context onto the 3D reconstruction.
  \item We find that the training of existing methods is undermined by misalignment between the geometric anchoring (rendering) and ground-truth videos. We thus introduce a camera re-localzation stage that reduces this alignment and notably improves synthesis accuracy.
  \item Extensive experiments on the challenging
    EgoExo4D~\citep{grauman2024egoexo4d} benchmark demonstrate that
    \textbf{Grounded-Exo2Ego} significantly outperforms recent
    state-of-the-art exo-to-ego generation methods across all metrics.
    Detailed ablations validate the contributions from our improvements at both the data-level and the architecture-level.
\end{itemize}

\section{Related work}
\label{sec:related}

\paragraph{Exo-to-ego video generation.}
We study single-exo-video to ego-video synthesis: given an exocentric video
and a target ego camera trajectory, the model must generate the
corresponding first-person video. Camera intrinsics and extrinsics are used
when available and estimated otherwise.
EgoExo4D~\citep{grauman2024egoexo4d} opened this line of work by providing
synchronized ego-exo recordings. Several follow-ups target related but
distinct settings: Exo2Ego-V~\citep{li2024exo2egov} and
Exo2EgoSyn~\citep{mahdi2025exo2egosyn} synthesize ego videos from
\emph{four} synchronized exo views -- Exo2EgoSyn
additionally injects pose-aware multi-view conditioning -- and
EgoWorld~\citep{park2025egoworld} addresses exo-to-ego \emph{image}
translation , not video. To our
knowledge, the only prior work in our exact single-exo-video to ego-video
setting is EgoX~\citep{egox2025}, which combines an ego-view geometric
prior and the exo video via width-wise latent
concatenation and uses geometry-guided
attention to encourage dense cross-view correspondence
.
EgoX relies primarily on geometry rendered from monocular depth, which
becomes unreliable wherever ego-view content is occluded or disoccluded in
the exo view. We keep the single-video setting and add two missing
ingredients: more reliable ego re-localization for the real-data pipeline,
and object-level semantic grounding that tells the generator what
corrupted or missing ego-view regions should depict.

\paragraph{Camera-controlled video and dynamic view synthesis.}
A second line of relevant work uses diffusion models for novel-view or
camera-controlled video generation. One thread conditions the generator on
\emph{point-cloud renderings} of the scene at the target view:
GEN3C~\citep{gen3c}, Uni3C~\citep{uni3c}, and Lyra~\citep{bahmani2026lyra}
lift the source view to a 3D point cloud and rasterize it at the target
camera as the conditioning image, while Vista4D~\citep{lin2026vista4d}
extends this design to dynamic 4D point clouds for re-shooting moving
scenes. A second thread conditions the generator on the
\emph{camera trajectory alone}, without an explicit 3D scaffold:
TrajectoryCrafter~\citep{yu2024trajectorycrafter},
ReCamMaster~\citep{bai2025recammaster},
SynCamMaster~\citep{bai2024syncammaster}, and Plenoptic Video
Generation~\citep{fu2025plenoptic} inject camera intrinsics and extrinsics
into the diffusion model, while PRoPE~\citep{li2025cameras} and
UCPE~\citep{zhang2025unified} reformulate camera control as a relative
positional encoding for video tokens. We compare against representatives
of both threads as baselines in \S\ref{sec:experiments}.

\section{Method}
\label{sec:method}

\subsection{Problem Setting}
\label{sec:problem_setting}

Single-video exocentric-to-egocentric (exo-to-ego) view synthesis takes as input an exocentric (exo)
video $\mathbf{V}^{\mathrm{exo}} \in \mathbb{R}^{F \times H_e \times W_e \times 3}$,
the exo camera intrinsics $\mathbf{K}^{\mathrm{exo}}$, the egocentric (ego)
camera intrinsics $\mathbf{K}^{\mathrm{ego}}$, and a target ego camera
trajectory $\{\mathbf{T}_f^{\mathrm{ego}}\}_{f=1}^{F}$.
The goal is to generate an egocentric video
$\hat{\mathbf{V}}^{\mathrm{ego}} \in \mathbb{R}^{F \times H \times W \times 3}$ consistent with the target trajectory. 

\subsection{Preliminaries}
\label{sec:diffusion_backbone}

We build upon a pre-trained latent video generative model~\citep{hacohen2026ltx2}. Let $\mathcal{E}$ and $\mathcal{D}$ denote the pre-trained VAE encoder and decoder.
We first obtain the clean latent of the ground-truth ego video
$\mathbf{z}_0 = \mathcal{E}(\bar{\mathbf{V}}^{\mathrm{ego}})$.
Following the flow-matching formulation~\citep{lipman2023flow}, we define a 
probability path between a standard Gaussian noise latent $\boldsymbol{\epsilon} \sim \mathcal{N}(\mathbf{0}, \mathbf{I})$ 
at $t=1$ and the data distribution at $t=0$. For a sampled time $t \in [0, 1]$, 
the interpolated latent is constructed as
$\mathbf{z}_t = t\mathbf{z}_0 + (1-t)\boldsymbol{\epsilon}.$
The target velocity field along this path is defined as the derivative with respect to time $\mathbf{u}_t = \mathbf{z}_0 - \boldsymbol{\epsilon}$.
The pre-trained Diffusion Transformer (DiT)~\citep{hacohen2026ltx2}
backbone, $f_\theta$, learns to predict this
velocity field conditioned on signals $\mathbf{c}$ by minimizing the objective
$
    \mathcal{L}_{\mathrm{FM}}
    = \mathbb{E}_{t,\mathbf{z}_0,\boldsymbol{\epsilon},\mathbf{c}}
    \bigl[
    \left\|
    f_\theta(\mathbf{z}_t, t, \mathbf{c}) - \mathbf{u}_t
    \right\|_2^2
    \bigl].
$
During inference,
we sample Gaussian noise $\mathbf{z}_{t=0} \sim \mathcal{N}(\mathbf{0}, \mathbf{I})$. We obtain 
the clean latent $\hat{\mathbf{z}}_0$ by continuously integrating the model predicted 
velocity field $f_\theta$ forward from $t=0$ to $t=1$. This generated latent is then 
decoded into the output video $\hat{\mathbf{V}}^{\mathrm{ego}} = \mathcal{D}(\hat{\mathbf{z}}_0)$.

\subsection{Overview}
\label{sec:method_overview}

\begin{figure}[t]
  \centering
  \includegraphics[width=\linewidth]{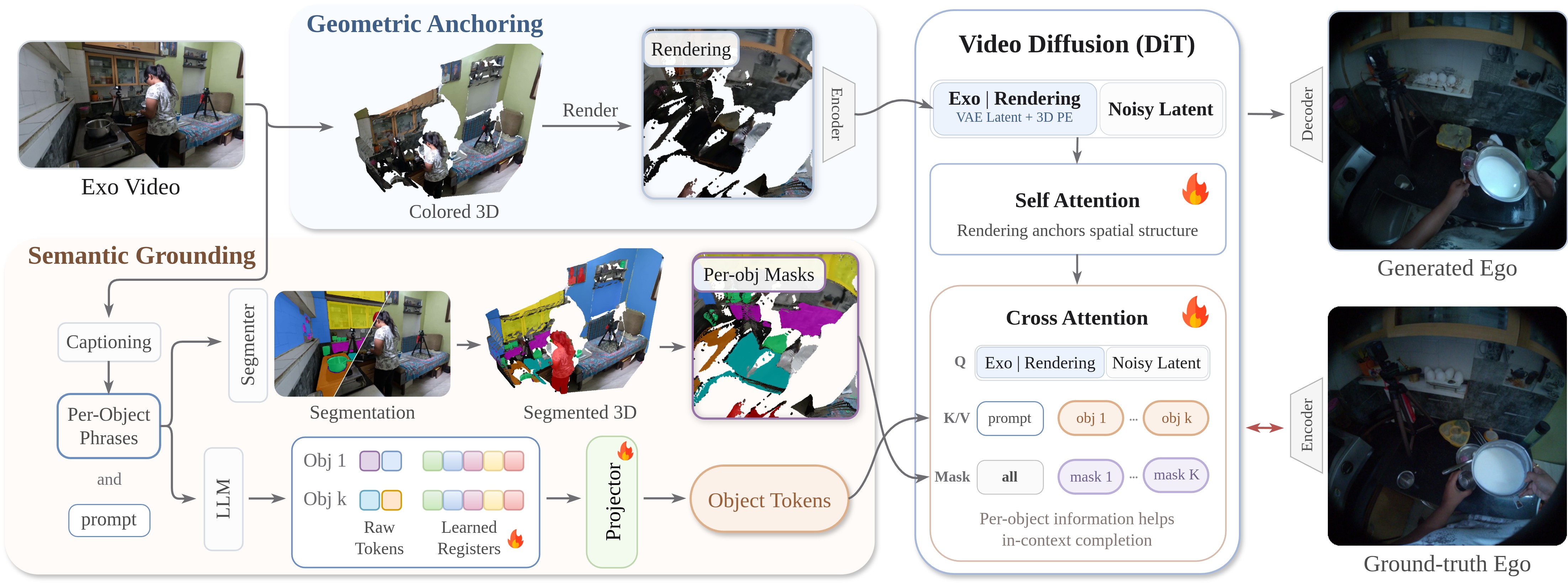}
  \caption{\textbf{Method overview.}
  \emph{Geometric anchoring} (top): Exocentric color and depth are lifted to a 3D
  reconstruction and rendered at the ego pose.
  \emph{Semantic grounding}
  (bottom): Per-object context is extracted as segmentation masks and object tokens. The segmentation masks are reprojected into the ego-view as used as attention masks during cross-attention. The per-object masks encourages noisy tokens to attend to  object information relevant to the specific region, providing spatial structure that grounds semantics onto pixels.}
  \label{fig:pipeline}
\end{figure}

Figure~\ref{fig:pipeline} presents an overview of our dual-branch
architecture for exo-to-ego video generation. 
The \emph{geometric anchoring} branch (\S\ref{sec:geometric_anchoring}) lifts
the exo view into 3D and renders it from the target ego camera, giving the
DiT a spatial scaffold for layout and camera motion. The \emph{semantic
grounding} branch (\S\ref{sec:semantic_grounding}) extracts object phrases
and masks from the exo view, reprojects the masks into the ego view, and
uses them to route object-specific text tokens to relevant video tokens during cross-attention. 
Both branches are
made reliable by our real and synthetic data pipelines (\S\ref{sec:data}).

\subsection{Geometric anchoring}
\label{sec:geometric_anchoring}

Geometric anchoring is the primary conditioning mechanism in state-of-the-art view-synthesis~\citep{lin2026vista4d,egox2025,gen3c,uni3c}. We adopt this strategy, reconstructing the scene in 3D, rendering it from the ego view, and injecting it into the DiT.
Given the exo-video depth and ego camera pose from our real-data pipeline (\S\ref{sec:data}),
we lift each exo pixel $\mathbf{p}=(u,v,1)^\top$ at frame $f$ into a 3D where $\mathbf{X}^{\mathrm{exo}}_f$:
\begin{equation}
  \mathbf{X}^{\mathrm{exo}}_f(u,v) = \mathbf{D}^{\mathrm{exo}}_f(u,v)\,(\mathbf{K}^{\mathrm{exo}})^{-1}\mathbf{p}.
  \label{eq:exo_lift}
\end{equation}
$\mathbf{D}_f^{\mathrm{exo}}$ is the depth at pixel $(u,v)$ in the $f$-th exo frame.
We then rasterize the 3D geometry at the egocentric camera using \texttt{nvdiffrast}~\citep{laine2020nvdiffrast}, yielding the egocentric rendering $\tilde{\mathbf{V}}^{\mathrm{ego}}_{f}$.
The pre-trained VAE, $\mathcal{E}$, encodes both videos into latent token sequences for the DiT:
\begin{equation}
  \mathbf{z}^{\mathrm{exo}} = \mathcal{E}(\mathbf{V}^{\mathrm{exo}}), \qquad \tilde{\mathbf{z}}^{\mathrm{ego}} = \mathcal{E}(\tilde{\mathbf{V}}^{\mathrm{ego}}).
  \label{eq:vae_encode}
\end{equation}
Here, $\tilde{\mathbf{z}}^{\mathrm{ego}}$ serves as the geometric anchor for the overall image structure.
To inject this condition, prior work~\citep{egox2025} channel-concatenates the render with the noisy latent $\mathbf{z}_{t}$.
This sometimes leads to large blank regions where the renderings are empty because invalid patches can become distracting signals despite the presence of an occupancy mask. 
Instead, following IC-LoRA~\citep{lhhuang2024iclora}, we append $\tilde{\mathbf{z}}^{\mathrm{ego}}$ as separate tokens and we mask out tokens from near-blank regions.
$\tilde{\mathbf{z}}^{\mathrm{ego}}_{\mathrm{masked}}$ uses the same $(f,x,y)$ RoPE~\citep{su2024rope} coordinates as the $\mathbf{z}_{t}$, thus preserving the geometric layout while safely deactivating unwanted near-blank tokens. Exocentric video tokens $\mathbf{z}^{\mathrm{exo}}$ are similarly concatenated, although their RoPE coordinates are shifted by the width of the ego video.
The model then generates the output ego video $\hat{\mathbf{V}}^{\mathrm{ego}}$ by iteratively denoising the noisy latent $\mathbf{z}_{t}$ as described in \S\ref{sec:diffusion_backbone}.

\subsection{Semantic grounding}
\label{sec:semantic_grounding}

Exo-to-ego view translation often requires extreme viewpoint changes, and geometric conditioning alone tends to produce errors when there are large distortions and holes. 
To synthesize plausible scenes from corrupted geometric signals, the model must know what kinds of objects to render and where to render them, \textit{i.e.} semantic information must be enriched with spatial structure.
We achieve structured semantic grounding by extracting rich semantic information from the exo-view and then use it to spatially guide text-to-video cross-attention.

\paragraph*{Extracting Rich Semantic Information.}
As shown in the bottom branch of Figure~\ref{fig:pipeline}, we deploy a VLM~\citep{xing2025caprl} to caption the scene into a descriptive scene prompt, including general caption, object description phrases, and spatial relationships. 
Then, the VLM deduplicates object candidates (for example, ``bike wheels'' and ``bicycle wheels'' become one candidate), and the surviving candidates are ranked by relevance to what the camera wearer is looking at and interacting with.
In practice, we select the top $K{=}10$ object candidates.
This cap is far less restrictive than it appears, for two reasons. First, each description is passed to the segmentation model as a text prompt, which returns all instances matching that phrase, so a single slot can cover a whole row of bike wheels rather than one wheel. The cap bounds the number of distinct object descriptions, not the number of segmented regions. Second, ten descriptions empirically cover most of the image area.

We then use an LLM~\citep{gemma2025gemma3} to encode the scene prompt and each per-object phrase into separate packets of text tokens, which are then projected to the latent space of the video DiT.  The $K$ token packets for the $K$ objects are each padded to a fixed length of $L{=}10$ tokens with a set of learned register tokens shared across objects. 
Each of the $K$ padded token packets is then transformed by a trainable projector into the latent space of the DiT cross-attention, producing the final token packet $\mathbf{O}_k$ for object $k$. 
The fixed length and shared learned registers give every object a stable key/value footprint regardless of phrase length.

\paragraph*{Semantic Grounding.} We reproject the exo segmentation mask $\mathbf{M}^{\mathrm{exo}}_k$ for object $k$ into the ego view to form $\mathbf{M}^{\mathrm{ego}}_k$ for frame $f$:
\begin{equation}
  \mathbf{M}^{\mathrm{ego}}_k(\mathbf{q})
  =
  \begin{cases}
    \mathbf{M}^{\mathrm{exo}}_k\!\left(\mathbf{U}_{f}(\mathbf{q})\right),
      & \text{if } \mathbf{D}_{f}(\mathbf{q})>0,\\
    0, & \text{otherwise}.
  \end{cases}
  \label{eq:mask_reproject}
\end{equation}

In the original text-to-video cross-attention layers, video tokens are used as queries and the scene's full caption prompt as key/value pairs. We enrich this mechanism by adding spatially structured attention patterns between per-object token packets $\{\mathbf{O}_k\}_{k=1}^{K}$ and the relevant video tokens to denoise (Figure~\ref{fig:pipeline}, right). 
Let $a_{n,k}\in\{0,1\}$ be the gating value for visual query token $n$ and object $k$. For an ego query token, $a_{n,k}$ comes from the reprojected ego mask $\mathbf{M}^{\mathrm{ego}}_k$, and for an exo query token from the exo mask $\mathbf{M}^{\mathrm{exo}}_k$. 
The cross-attention score from query token $n$ to
object-token $(k,m)$ is gated by $a_{n,k}$:
\begin{equation}
  A^{\mathrm{obj}}_{n,(k,m)}
  =
  \begin{cases}
    0, & a_{n,k}=1,\\
    -\infty, & a_{n,k}=0.
  \end{cases}
  \label{eq:object_gate}
\end{equation}
Only visual tokens inside object $k$'s region attends to $k$'s packet of text tokens, and every other token is blocked. 
In this way, the gating masks facilitate spatial grounding by allowing object semantics to stay in the original text latent space of the pretrained backbone. We also explored other approaches such as injecting DINOv3 features~\citep{simeoni2025dinov3}), directly spatializing the object tokens into feature maps, etc. We found these approaches to be far less effective compared to our design.

Due to practical limits on compute and storage, we only compute segmentation masks for the first frame and broadcast the same mask across time, i.e., $f=0$ in Eq.~\ref{eq:mask_reproject}.
We observe that this approach already yields strong improvement (Table~\ref{tab:main}). Full video segmentation might yield further gains at higher cost.
Additionally, during this grounding step, cross-attention to the full scene prompt is unmodified, \textit{i.e.} all visual tokens still attend to the scene prompt.

\subsection{Real and Synthetic Data Pipelines}
\label{sec:data}

As discussed in the Introduction~\S\ref{sec:intro}, monocular reconstructions are misaligned to real-world geometry. As a result, ground-truth ego poses (calibrated to real-world geometry) are incompatible with monocular reconstructions.
Therefore, directly rendering the reconstructions using ground-truth pose yields videos that are misaligned to the ground-truth ego video, severely undermining the learning progress. 
Providing well-aligned renderings during test time does not meaningfully improve quality because the models have already learned to distrust the misaligned renderings during training.
We address this issue through two data contributions: (1) camera re-localization, and (2) a generative synthetic data engine.

\paragraph{Ego camera re-localization for real data.}
In order to improve the alignment between the ego rendering
$\tilde{\mathbf{V}}^{\mathrm{ego}}$ and ground-truth $\bar{\mathbf{V}}^{\mathrm{ego}}$ for better learning,
ego cameras must be aligned to the monocular reconstruction.
One approach is to directly estimate the person's head pose using human pose estimation or face trackers. However, this is highly unreliable due to occlusion and challenging head motion. 
Instead, we refine the provided ego trajectory labels with respect to the monocular reconstruction as follows:
(1) We first generate monocular reconstructions using MoGe2~\citep{wang2025moge2} on the exo video;
(2) then we use Depth-Anything-3(DA3)~\citep{lin2025depthanything3} to create a metric-calibrated reconstruction using all ground-truth exo camera as input;
(3) We non-rigidly align the DA3 depth map to the MoGe2 depth, yielding a transformation from the real-world to MoGe2's coordinate frame;
(4) We then apply this non-rigid alignment to the ground-truth ego pose to bring it into MoGe2's monocular reconstruction. During this process, the original ground-truth camera rotations remain unchanged, and only the depth of the ego camera is modified (and thus the camera center in 3D).

Notice that this step establishes \emph{coordinate consistency} rather than
\emph{geometric accuracy}. No choice of camera pose can turn a locally incorrect
depth estimate into correct geometry. Instead, what it resolves is the challenge of finding a camera pose within the MoGe2 reconstruction that would produce a rendering close to the ground-truth ego video. 
Additionally, we deliberately use MoGe2's original reconstructions, which is inaccurate, such that the generator learns to complete and correct the rendering under the same class of error it will encounter at inference.
Otherwise, training with perfect reconstructions would introduce a train/test mismatch, thus undermining the model's inference performance when it encounters imperfect reconstructions. 
Note that DA3 serves only as a coordinate bridge for deriving the camera label and is never an input to the model.
In Table~\ref{tab:reloc-shift} (Appendix~\ref{app:reconstruction}), we share statics about the magnitude of such adjustment. For example, ego cameras with depth values larger than 6 meters are adjusted with a median amount of 1.86 meters. Table~\ref{tab:ablation} shows that this camera re-localization stage leads to substantial improvements across all metrics.

\paragraph{Automated synthetic data engine.}
The synthetic engine produces exo/ego pairs with accurately labeled ground-truth, 
cleanly separating the learning of semantic grounding 
from the challenge of staying robust to corrupted geometry. 
We generate characters via a customized character
pipeline that leverages \textit{gpt-image-1} and \textit{Meshy.ai} to sequentially generate character images from text prompts, and 3D characters from the images. We further develop a custom algorithm to reliably rig the generated 3D characters, and animate them with real interaction motion sequences from Embody3D~\citep{embody3d}.
We generate interior scenes with Infinigen~\citep{raistrick2023infinigen}.
We automatically generate an asset pool of 640 interior scenes, 550 rigged characters, and over 20K training clips.
Appendix~\ref{app:syntheticdataset} describes the
full asset, rigging, camera, and rendering pipeline.

\section{Experiments}
\label{sec:experiments}

\subsection{Experimental setup}
\label{sec:setup}

\paragraph{Dataset.}
We follow EgoX~\citep{egox2025} to train and evaluate on EgoExo4D~\citep{grauman2024egoexo4d} using the \emph{Unseen} (never seen during training) and \emph{New Action} splits (renamed from ``Seen'' in EgoX, new actions but in environments seen during training).
EgoExo4D is a standard benchmark with in-the-wild corpus captured across several countries rather than a studio. It spans large and fast body motion, a wide range of tasks (such as bike repair, cooking, soccer, and health procedures), and diverse environments (such as homes, sports fields, parking lots, and conference rooms). This in-the-wild benchmark thus evaluates methods under imperfect geometry, segmentation masks, and camera poses.

\begin{table}[t]
  \caption{\textbf{Quantitative comparison on EgoExo4D.} Best in
  \textbf{bold}, second-best \underline{underlined}. Evaluated on the samples using EgoX's evaluation protocol. 
  \emph{Unseen} indicates brand new environments not seen during training.
  \emph{New Action} (renamed from EgoX's \emph{Seen}) indicate unseen actions but in environment seen during training.
  Grey rows show the change over EgoX.
  Our numbers are averaged over five independent generations (standard
  deviations in Appendix~\ref{app:seedvar}).
  \emph{Loc Err} and \emph{Contour} are normalized by the output side length. Values in \underline{[brackets]} are metrics the EgoX paper does not
  report so we measured them ourselves.}
  \label{tab:main}
  \centering
  \setlength{\tabcolsep}{3pt}
  \resizebox{\textwidth}{!}{%
  \begin{tabular}{ll cccc ccc cc}
    \toprule
    & & \multicolumn{4}{c}{\textbf{Image Metrics}}
    & \multicolumn{3}{c}{\textbf{Object Metrics}}
    & \multicolumn{2}{c}{\textbf{Video Metrics}} \\
    \cmidrule(lr){3-6} \cmidrule(lr){7-9} \cmidrule(lr){10-11}
    & Method
    & PSNR$\uparrow$ & SSIM$\uparrow$ & LPIPS$\downarrow$ & CLIP-I$\uparrow$
    & Loc Err$\downarrow$ & IoU$\uparrow$ & Contour$\downarrow$
    & FVD$\downarrow$ & T-LPIPS$\downarrow$ \\
    \midrule
    \multirow{8}{*}{\rotatebox{90}{\shortstack{New Action\\[1pt]{\scriptsize(new actions, seen envs.)}}}}
    & Exo2Ego-V~\citep{li2024exo2egov}        & 14.53 & 0.384 & 0.569 & 0.774 & 0.326 & 0.074 & \textcolor{gray}{--} & 622.47  & \textcolor{gray}{--} \\
    & TrajectoryCrafter~\citep{yu2024trajectorycrafter} & 13.05 & 0.375 & 0.606 & 0.780 & 0.210 & 0.128 & \textcolor{gray}{--} & 546.09  & \textcolor{gray}{--} \\
    & Wan-Fun-Control~\citep{wanfuncontrol2025}   & 12.25 & 0.463 & 0.617 & 0.810 & 0.235 & 0.076 & \textcolor{gray}{--} & 595.07  & \textcolor{gray}{--} \\
    & Wan-VACE~\citep{jiang2025vace}          & 12.95 & 0.413 & 0.626 & 0.829 & 0.228 & 0.114 & \textcolor{gray}{--} & 508.69  & \textcolor{gray}{--} \\
    & Vista4D~\citep{lin2026vista4d}          & 10.39 & 0.301 & 0.670 & 0.796 & 0.139 & 0.254 & 0.090 & 647.51 & 4.40 \\
    & EgoX~\citep{egox2025}              & \underline{16.05} & \underline{0.556} & \underline{0.498} & \underline{0.896} & \underline{0.129} & \underline{0.363} & \underline{[0.071]} & \underline{184.47} & \underline{[2.688]} \\
    & \textbf{Ours}     & \textbf{18.64} & \textbf{0.560} & \textbf{0.346} & \textbf{0.910} & \textbf{0.031} & \textbf{0.576} & \textbf{0.027} & \textbf{119.9} & \textbf{1.454} \\
    \rowcolor{gray!15}
    & \hspace{1em}$\Delta$ vs EgoX & $+2.59$\,dB & $+0.7\%$ & $-30.5\%$ & $+1.6\%$ & $-76.0\%$ & $+58.7\%$ & $-62.0\%$ & $-35.0\%$ & $-45.9\%$ \\
    \midrule
    \multirow{8}{*}{\rotatebox{90}{\shortstack{Unseen\\[1pt]{\scriptsize(new actions, new envs.)}}}}
    & Exo2Ego-V~\citep{li2024exo2egov}        & 12.70 & 0.439 & 0.597 & 0.679 & 0.447 & 0.003 & \textcolor{gray}{--} & 1283.50 & \textcolor{gray}{--} \\
    & TrajectoryCrafter~\citep{yu2024trajectorycrafter} & 12.24 & 0.297 & 0.619 & 0.778 & 0.400 & 0.039 & \textcolor{gray}{--} & 821.71  & \textcolor{gray}{--} \\
    & Wan-Fun-Control~\citep{wanfuncontrol2025}   & 13.59 & 0.439 & 0.604 & 0.799 & 0.399 & 0.042 & \textcolor{gray}{--} & 968.78  & \textcolor{gray}{--} \\
    & Wan-VACE~\citep{jiang2025vace}          & 12.17 & 0.345 & 0.638 & 0.820 & 0.400 & 0.038 & \textcolor{gray}{--} & 1045.45 & \textcolor{gray}{--} \\
    & Vista4D~\citep{lin2026vista4d}          & 10.68 & 0.255 & 0.668 & 0.811 & \underline{0.146} & \underline{0.229} & 0.087 & 794.09 & 2.91 \\
    & EgoX~\citep{egox2025}              & \underline{14.38} & \underline{0.457} & \underline{0.552} & \underline{0.877} & 0.312 & 0.092 & \underline{[0.083]} & \underline{440.64} & \underline{[2.375]} \\
    & \textbf{Ours}     & \textbf{16.05} & \textbf{0.460} & \textbf{0.467} & \textbf{0.894} & \textbf{0.052} & \textbf{0.412} & \textbf{0.044} & \textbf{385.8} & \textbf{1.485} \\
    \rowcolor{gray!15}
    & \hspace{1em}$\Delta$ vs EgoX & $+1.67$\,dB & $+0.7\%$ & $-15.4\%$ & $+1.9\%$ & $-83.3\%$ & $+347.8\%$ & $-47.0\%$ & $-12.4\%$ & $-37.5\%$ \\
    \bottomrule
  \end{tabular}%
  }
\end{table}

\begin{figure}[t]
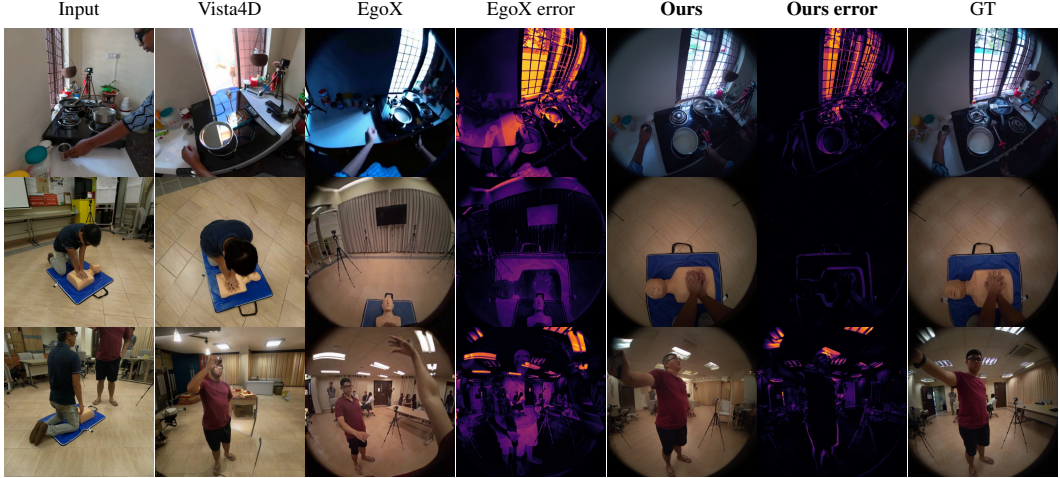

  \comparisonfigureseenA
  \caption{\textbf{Qualitative comparison on EgoExo4D (New Activities).} }
  \label{fig:comparison-seen}
\end{figure}

\begin{table}[t]
  \caption{\textbf{Ablation studies on EgoExo4D.} 
  \textsc{Top}: the components each
  variant uses. \textsc{Bottom}: results, averaged over five independent generations
  (standard deviations in Appendix~\ref{app:seedvar}). 
  \emph{Loc Err} and \emph{Contour} are normalized by the
  image side length. Bracketed EgoX values are the two metrics its paper does not report, which we measured ourselves.
  \textbf{Rows marked $^{*}$ are our own re-implementations of EgoX}, ported to
  the LTX-2.3 backbone so that the comparison isolates one component at a time;
  EgoX's released code targets a different backbone, so these were rebuilt to
  the best of our ability from the paper and the reference implementation, and
  they may understate what the original authors would achieve. The unmarked
  \emph{EgoX} row carries the numbers reported in their paper.}
  \label{tab:ablation}
  \centering
  \setlength{\tabcolsep}{3pt}
  \small
  \resizebox{\textwidth}{!}{%
  \begin{tabular}{l ccccccc}
    \toprule
    Variant & Backbone & Our data & Camera reloc. & Seq. concat & Obj tok. & Obj mask & Synth data \\
    \midrule
    EgoX~\citep{egox2025}   & Wan & -- & -- & -- & -- & -- & -- \\
    EgoX (LTX, data)$^{*}$        & LTX & Y  & -- & -- & -- & -- & -- \\
    EgoX (LTX, data, reloc)$^{*}$ & LTX & Y  & Y  & -- & -- & -- & -- \\
    Ours (no obj tok, mask, synth) & LTX & Y  & Y  & Y  & -- & -- & -- \\
    Ours (no obj mask, synth)   & LTX & Y  & Y  & Y  & Y  & -- & -- \\
    Ours (no synth)         & LTX & Y  & Y  & Y  & Y  & Y  & -- \\
    \textbf{Ours}           & LTX & Y  & Y  & Y  & Y  & Y  & Y  \\
    \bottomrule
  \end{tabular}%
  }

  \vspace{1.2em}

  \setlength{\tabcolsep}{3pt}
  \resizebox{\textwidth}{!}{%
  \begin{tabular}{ll cccc ccc cc}
    \toprule
    & & \multicolumn{4}{c}{\textbf{Image Metrics}}
    & \multicolumn{3}{c}{\textbf{Object Metrics}}
    & \multicolumn{2}{c}{\textbf{Video Metrics}} \\
    \cmidrule(lr){3-6} \cmidrule(lr){7-9} \cmidrule(lr){10-11}
    & Variant
    & PSNR$\uparrow$ & SSIM$\uparrow$ & LPIPS$\downarrow$ & CLIP-I$\uparrow$
    & Loc Err$\downarrow$ & IoU$\uparrow$ & Contour$\downarrow$
    & FVD$\downarrow$ & T-LPIPS$\downarrow$ \\
    \midrule
    \multirow{8}{*}{\rotatebox{90}{\shortstack{New Action\\[1pt]{\scriptsize(new actions, seen envs.)}}}}
    & EgoX~\citep{egox2025}   & 16.05 & 0.556 & 0.498 & 0.896 & 0.129 & 0.363 & [0.071] & 184.47 & [2.688] \\
    & EgoX (LTX, data)$^{*}$        & 17.05 & 0.506 & 0.473 & 0.875 & 0.065 & 0.336 & 0.061 & 233.4 & 3.115 \\
    & EgoX (LTX, data, reloc)$^{*}$ & 17.95 & 0.532 & 0.403 & 0.889 & 0.045 & 0.456 & 0.041 & 169.6 & 2.354 \\
    & Ours (no obj tok, mask, synth) & 17.69 & 0.529 & 0.400 & 0.894 & 0.040 & 0.501 & 0.036 & 163.9 & 1.593 \\
    & Ours (no obj mask, synth)   & 18.06 & 0.541 & 0.377 & 0.900 & 0.035 & 0.540 & 0.031 & 142.8 & 1.540 \\
    & Ours (no synth)         & \underline{18.39} & \underline{0.552} & \underline{0.355} & \underline{0.908} & \underline{0.032} & \underline{0.569} & \underline{0.028} & \underline{122.9} & \textbf{1.417} \\
    & \textbf{Ours}           & \textbf{18.64} & \textbf{0.560} & \textbf{0.346} & \textbf{0.910} & \textbf{0.031} & \textbf{0.576} & \textbf{0.027} & \textbf{119.9} & \underline{1.454} \\
    \cmidrule(lr){2-11}
    & \emph{Given ego attn mask} & \emph{18.94} & \emph{0.568} & \emph{0.330} & \emph{0.913} & \emph{0.026} & \emph{0.605} & \emph{0.022} & \emph{113.8} & \emph{1.437} \\
    \midrule
    \multirow{8}{*}{\rotatebox{90}{\shortstack{Unseen\\[1pt]{\scriptsize(new actions, new envs.)}}}}
    & EgoX~\citep{egox2025}   & 14.38 & 0.457 & 0.552 & 0.877 & 0.312 & 0.092 & [0.083] & 440.64 & [2.375] \\
    & EgoX (LTX, data)$^{*}$        & 15.50 & 0.439 & 0.533 & 0.853 & 0.085 & 0.243 & 0.078 & 534.6 & 2.693 \\
    & EgoX (LTX, data, reloc)$^{*}$ & 15.92 & 0.449 & 0.497 & 0.868 & 0.070 & 0.316 & 0.063 & 467.4 & 2.063 \\
    & Ours (no obj tok, mask, synth) & 15.67 & 0.445 & 0.491 & 0.882 & 0.062 & 0.374 & 0.054 & 417.9 & 1.598 \\
    & Ours (no obj mask, synth)   & 15.76 & 0.451 & 0.481 & 0.883 & 0.058 & 0.385 & 0.050 & 399.9 & 1.596 \\
    & Ours (no synth)         & \underline{16.00} & \textbf{0.460} & \textbf{0.462} & \underline{0.892} & \textbf{0.052} & \textbf{0.418} & \underline{0.046} & \textbf{372.4} & \textbf{1.416} \\
    & \textbf{Ours}           & \textbf{16.05} & \textbf{0.460} & \underline{0.467} & \textbf{0.894} & \textbf{0.052} & \underline{0.412} & \textbf{0.044} & \underline{385.8} & \underline{1.485} \\
    \cmidrule(lr){2-11}
    & \emph{Given ego attn mask} & \emph{16.30} & \emph{0.467} & \emph{0.453} & \emph{0.897} & \emph{0.046} & \emph{0.451} & \emph{0.038} & \emph{373.6} & \emph{1.462} \\
    \bottomrule
  \end{tabular}%
  }
\end{table}

\begin{figure}[t]
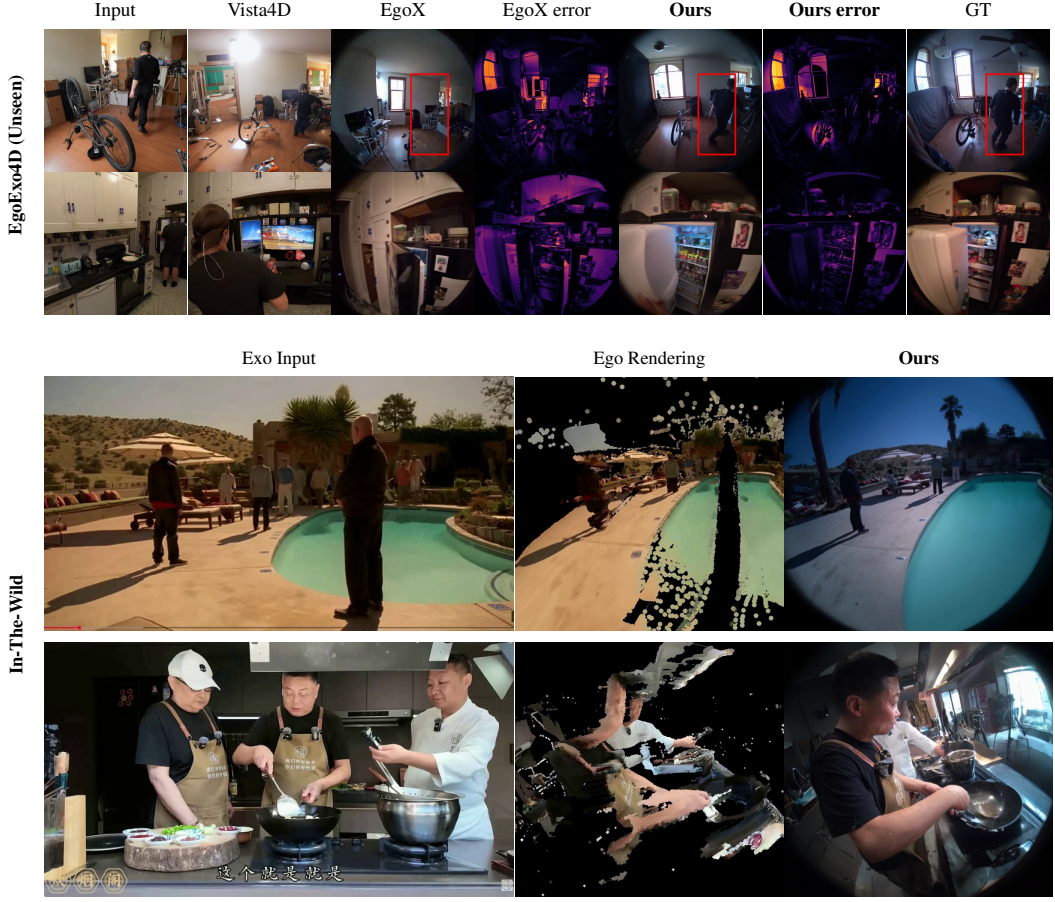

  \comparisonfigureunseen
  \caption{Qualitative comparisons on EgoExo4D (top) and in-the-wild samples (bottom).
  }
  \label{fig:comparison-unseen}
\end{figure}

\begin{figure}[t]
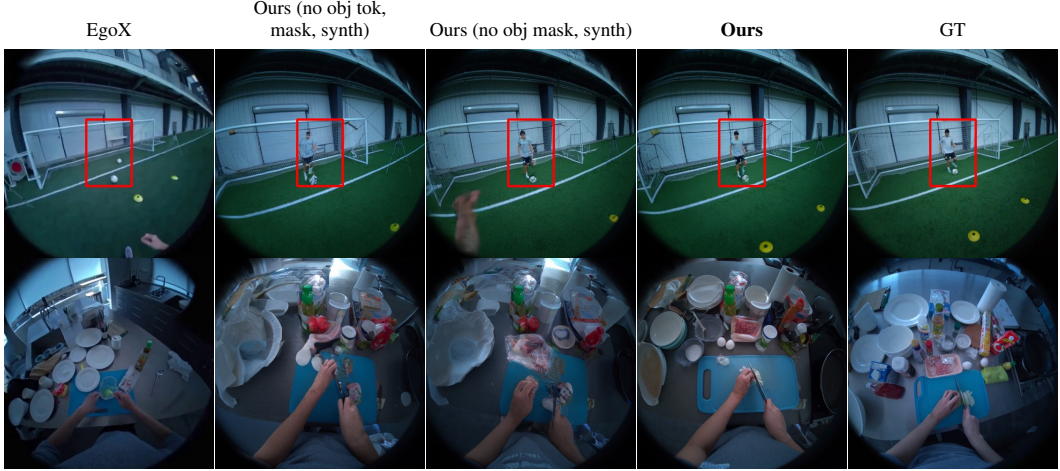

  \comparisonfigureablation
  \caption{\textbf{Ablation study.} Top row: EgoX omits the player entirely, but our variants reconstruct the player progressively closer to the ground truth. Bottom row: EgoX fails to reconstruct the viewpoint and scene correctly, but our variants reconstruct the ego view progressively more accurately.}
  \label{fig:comparison-ablation}
\end{figure}

\paragraph{Implementation details.}
We use LTX-2.3~\citep{hacohen2026ltx2}
as our base video diffusion model with 256 trainable LoRA ranks on self-attention and cross-attention layers. 
We train on 64 NVIDIA H100 GPUs with mini-batches of 1 on each GPU, learning rate $5{\times}10^{-5}$, and half precision. We use $F{=}33$, $H{=}448$, $W{=}448$, and $K{\leq}10$ objects per clip, each carrying $L{=}10$ tokens.

\paragraph{Evaluation metrics.}
We report three groups, similar to EgoX: \textbf{Image
Metrics} (PSNR, SSIM, LPIPS, CLIP-I), \textbf{Video Metrics}
(FVD~\citep{DBLP:journals/corr/abs-1812-01717} and Temporal-LPIPS
(T-LPIPS)~\citep{chu2020tlpips}), and \textbf{Object Metrics}
(Location Error, IoU, and Contour Chamfer distance).
Object Metrics are computed over object pairs matched using EgoX's protocol using SAM2-Large~\citep{ravi2024sam2} and
DINOv3~\citep{simeoni2025dinov3} for matching.
EgoX's Contour Accuracy protocol is no longer available, so we choose Chamfer distance normalized by image resolution as an easily reproducible replacement.
The three VBench~\citep{zhang2024vbench} metrics are deferred to Appendix~\ref{app:vbench} because these no-reference metrics do not measure the similarity between generated and ground-truth videos. Full definitions and additional metrics are in
Appendix~\ref{app:metrics}.

\paragraph{Baselines.}
We compare against EgoX~\citep{egox2025} as well as its baselines (Exo2Ego-V~\citep{li2024exo2egov},
TrajectoryCrafter~\citep{yu2024trajectorycrafter},
Wan-Fun-Control~\citep{wanfuncontrol2025},
Wan-VACE~\citep{jiang2025vace})
with numbers taken from EgoX~\citep{egox2025}.
Additionally, we adapt the recent state-of-the-art method in general view synthesis, Vista4D~\citep{lin2026vista4d}, to our setting to demonstrate the need for dedicated exo-to-ego methods. 
For a fair comparison, we also perform camera re-localization by aligning its Pi3X~\citep{wang2025pi} reconstruction to ours. 
Two concurrent works, Exo2EgoSyn~\citep{mahdi2025exo2egosyn} (multiview-to-ego) and EgoWorld~\citep{park2025egoworld} (image-only), are not directly comparable, and neither has released code at the time of writing, so we
exclude them from Table~\ref{tab:main}.

\subsection{Comparisons}
\label{sec:main_results}

\paragraph{Quantitative comparisons.}
As shown in Table~\ref{tab:main}, our method substantially outperforms EgoX and recent state-of-the-art methods by a wide margin.
The significant improvements in metrics such as PSNR ($+1.67$\,dB$/+2.59$\,dB on \textit{Unseen\,/\,New~Action}) and LPIPS ($-15.4\%/-30.5\%$) indicate substantial improvements in reconstruction quality and alignment accuracy.
The large improvement in T-LPIPS ($-37.5\%/-45.9\%$) indicates that our model reconstructs temporal dynamics (content dynamics and camera motion) much better than recent state-of-the-art methods.
The drastic improvements in Object Metrics (Loc Err: $-83.3\%/-76.0\%$; IoU: $+347.8\%/+58.7\%$; Contour: $-47.0\%/-62.0\%$) indicate that not only are the video's general structure improved significantly, but object-level generation and localization are also substantially improved.
The large improvements in generative metrics such as FVD ($-12.4\%/-35.0\%$) prove that our method generates videos much more similar to real egocentric videos in visual characteristics than existing methods.

\paragraph{Qualitative comparisons.}
As seen from Fig.~\ref{fig:comparison-seen}, Vista4D consistently fails to generate ego views correctly. While EgoX performs better, its results show significant misalignment against the ground-truth. Our results consistently reconstruct the scene with accurate spatial alignment as well as the content of the scene. For example, the red boxes in Fig.~\ref{fig:comparison-unseen} and ~\ref{fig:comparison-ablation} show that EgoX completely fails to reconstruct the persons, whereas our method succeeds.

The significant improvement validates our hypothesis that
(1) dedicated exo-to-ego is needed as shown by the gap between Vista4D and EgoX, and (2) exo-to-ego benefits from a systematic solution spanning the data pipeline and the architecture. Notably, by adding semantics on top of the geometric scaffold, our method moves past the geometry-centric design that state-of-the-art methods share and achieves further improvements. Additionally, despite the significant improvements, the absolute performance still has much room to improve, for example, the average PSNR is still under 20dB.

\subsection{Ablation studies}
\label{sec:ablations}
Table~\ref{tab:ablation} shows ablation studies on each of our proposed contributions, together with the components each variant uses.
The first three variants progress from EgoX to our ungrounded baseline, so the credit due to the backbone and more training data, and the camera re-localization can each be separated from semantic grounding's contributions. All deltas are quoted as Unseen\,/\,New~Action. Figure~\ref{fig:comparison-ablation} shows the corresponding qualitative comparisons.

\textbf{Improved backbone and training data.}
\emph{EgoX (LTX, data)}$^{*}$ --- our re-implementation of EgoX, rebuilt to the best of our ability --- changes the backbone from Wan2.1 $\to$ LTX-2.3 and uses our training data, which is a subset of EgoEgo4D's training split but larger than EgoX. EgoX discards a large share of EgoExo4D because misalignment between the ego camera and the monocular reconstruction leaves the rendered prior mostly blank. Our pipeline keeps those takes. These two changes lift reconstruction fidelity (PSNR $+1.12/+1.00$~dB, LPIPS $-3.4\%/-5.0\%$, Loc Err $-72.8\%/-49.6\%$) but leaves the video metrics \emph{worse} than EgoX (FVD $+21.3\%/+26.5\%$, T-LPIPS $+13.4\%/+15.9\%$).

\textbf{Camera re-localization.}
\emph{EgoX (LTX, data, reloc)}$^{*}$, again our own re-implementation, further adds our ego-camera re-localization, which aligns the ego cameras to the monocular reconstruction and thereby better aligns the rendered ego prior with the ground-truth ego video. All nine metrics improve, the geometric ones most: IoU $+30.0\%/+35.7\%$, Contour error $-19.2\%/-32.8\%$, PSNR $+0.42/+0.90$~dB, LPIPS error $-6.8\%/-14.8\%$, FVD error $-12.6\%/-27.3\%$. The synthesized videos become not only more spatially but also more temporally aligned with the ground truth (T-LPIPS error $-23.4\%/-24.4\%$). This matches what we observe qualitatively: EgoX often does not respect its rendering conditioning, because that conditioning is rarely reliable during training and the model learns to distrust it.

\textbf{Sequence-concatenated versus channel-concatenated conditioning}.
\emph{Ours (no obj tok, mask, synth)} and \emph{EgoX (LTX, data, reloc)}$^{*}$ (our re-implementation) are similar geometry-only variants on the high-level. They are both conditioned on the rendered video and exo video, and differ only in the conditioning mechanism: we concatenate it along the sequence dimension (IC-LoRA~\cite{lhhuang2024iclora} style), whereas every EgoX variant concatenates it along the channel dimension of the noisy video tokens. The trade is informative. PSNR falls slightly ($-0.25/-0.26$~dB) while structure improves clearly (IoU $+18.4\%/+9.9\%$, Contour error $-14.3\%/-12.2\%$, T-LPIPS error $-22.5\%/-32.3\%$, FVD error $-10.6\%/-3.4\%$): channel concatenation optimizes the per-pixel mean, sequence concatenation preserves the layout.

\textbf{Object tokens without grounding.}
\emph{Ours (no obj mask, synth)} adds per-object semantic tokens on top of the rich caption every variant already receives. These per-object tokens are globally available to all video tokens. Gains are broad though modest: IoU $+2.9\%/+7.8\%$, Contour error $-7.4\%/-13.9\%$, PSNR $+0.09/+0.37$~dB, FVD $-4.3\%/-12.9\%$. Dedicated object tokens are therefore useful in their own right.

\textbf{Full semantic grounding.}
Comparing to \emph{Ours (no obj tok, mask, synth)} which does not utilize object tokens or segmentation masks, \emph{Ours} improves IoU by $+10.2\%/+15.0\%$, Contour error by $-18.5\%/-25.0\%$, PSNR by $+0.38/+0.95$~dB, LPIPS error by $-4.9\%/-13.5\%$, T-LPIPS error by $-7.1\%/-8.7\%$, and FVD error by $-7.7\%/-26.8\%$. Full semantic grounding therefore improves both the scene-level and object-level accuracy.

\textbf{Spatial mask is important for grounding semantics.}
Comparing to \emph{Ours (no obj mask, synth)}, \emph{Ours} gains most on the axes the gate acts on: Loc Err $-10.3\%/-11.4\%$, Contour error $-12.0\%/-12.9\%$, and IoU $+7.0\%/+6.7\%$, followed by T-LPIPS error $-7.0\%/-5.6\%$ and FVD error $-3.5\%/-16.0\%$, with PSNR last at $+0.29/+0.58$~dB. Holding the training data fixed (\emph{Ours (no obj mask, synth)} vs.\ \emph{Ours (no synth)}) isolates the mask on its own and preserves the same ordering: Loc Err $-10.3\%/-8.6\%$, Contour error $-8.0\%/-9.7\%$, IoU $+8.6\%/+5.4\%$. The spatial gate supplies location-conditioned routing that improves alignment.

\textbf{Synthetic data lifts in-distribution quality.}
Comparing to \emph{Ours (no synth)} which does not use the synthetic data, \emph{Ours} improve PSNR by $+0.25$~dB for the \emph{New Action} test set. 

\textbf{Accurate segmentation masks improve quality.}
We evaluate the how mask accuracy impacts quality by testing with segmentation mask obtained from the first frame of the actual ego video (\emph{given ego attn mask}) instead of using reprojected masks. This improves every metric --- PSNR by $+0.25/+0.30$~dB, IoU by $+9.5\%/+5.0\%$, Loc Err by $-11.5\%/-16.1\%$, and Contour by $-13.6\%/-18.5\%$. Although this is not deployable at test time, it confirms that improved structural semantic grounding can improve quality. One interesting direction is to generate semantic masks as intermediate outputs as part of visual-chain-of-thoughts. Our early experiments show that an additional Mask2Former-style~\citep{mask2former} branch
can generate rather accurate semantic masks, and we leave this for future work.

\subsection{Limitations.}
\textbf{Long video generation.} The video length is limited due to VRAM constraints. Extending to longer videos could be done through chunk-by-chunk denoising as explored by recent methods. On our website, we share our on-going work which shows that a short and simple chunk-autoregressive finetuning stage allows the generation of longer videos (e.g. 10 seconds) with stable quality.

\textbf{Occlusion, out-of-frame content, and dependence on 3D reconstruction.} Our model utilizes off-the-shelf 3D reconstruction methods for video generation. While 3D reconstruction methods have been improving rapidly in terms of accuracy and completeness, a persistent challenge is that occlusion and out-of-frame contents are difficult to reconstruct. When the reconstruction is largely blank in the view of the ego camera, the model needs to hallucinate most of the content so accuracy can thus be undermined. When the exo camera is allowed to move across the scene to gather information from different viewpoints, this challenge can be alleviated, though hard to eliminate. This is a fundamental challenge, and we leave it for future work. Appendix~\ref{app:robust} quantifies how far this dependence extends.

\section{Conclusion}
\label{sec:conclusion}

We proposed a framework for exo-to-ego video generation that tackles the problem at both the architectural and the data levels. First, we build a solid foundation for this problem by resolving inherent limitations in existing data pipelines and also introducing a generative data engine, leading to significant improvement. More importantly, we demonstrate a novel Semantic-Grounding mechanism that explores beyond traditional geometry-based video synthesis and further improves quality by guiding the generation with spatially anchored semantics.  
Ablation studies demonstrates the effectiveness of our various proposed contributions.

\bibliographystyle{plainnat}
\bibliography{references}

\newpage
\appendix
\section*{Supplementary Material}

\section{Additional qualitative results}
\label{app:additional}

\begin{figure}[!htbp]
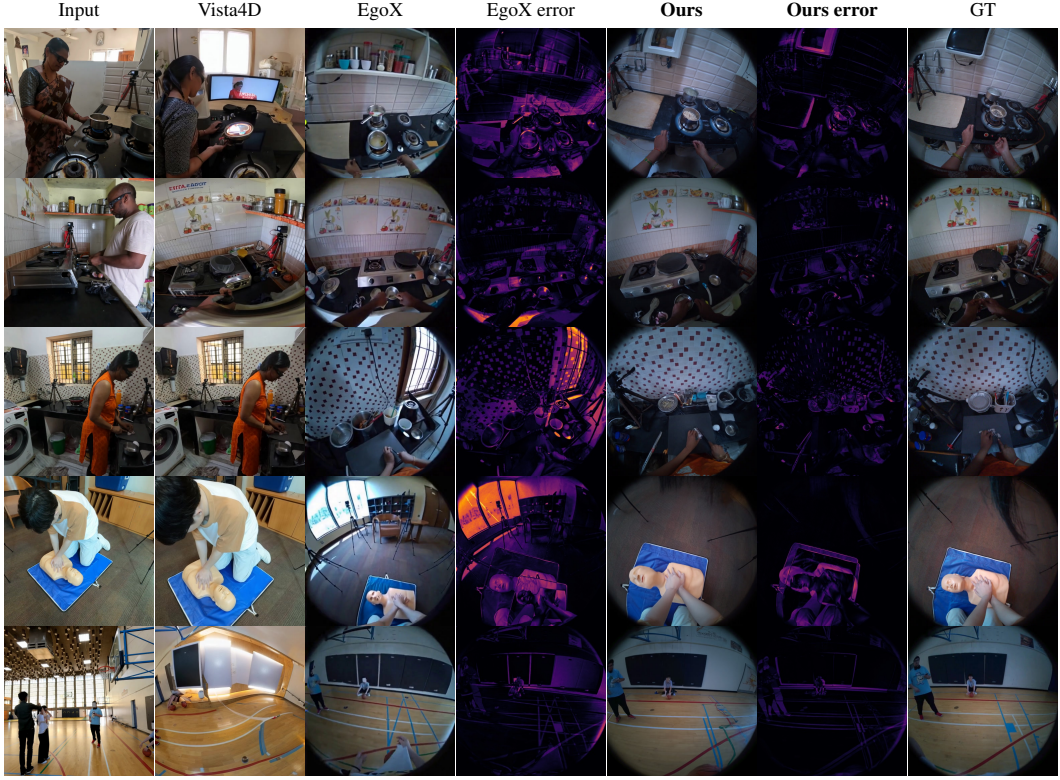

  \comparisonfigureseensupplemental
  \caption{\textbf{Additional New-Activity scenes.} Columns as in Figure~\ref{fig:comparison-seen}.}
  \label{fig:comparison-seen-supp}
\end{figure}

\section{Evaluation metrics}
\label{app:metrics}

This section gives the definitions for the Chamfer contour distance and describes why we deferred the VBench metrics.

\paragraph{Chamfer Contour distance.}
Following EgoX's convention, we run SAM2-Large~\citep{ravi2024sam2}
to automatically generate segmentation masks for each of the generated and GT frames. We embed each detected object using DINOv3 ViT-L/16~\citep{simeoni2025dinov3}. We then match objects based on their cosine similarities of DINOv3 features and Hungarian matching (this is shared with the
Location Error and IoU metrics as done in EgoX). 
For each pair of masks we extract the boundary pixel set
$\mathcal{B}^{\text{gen}}$ and $\mathcal{B}^{\text{gt}}$ and compute the
two-sided Chamfer distance
\[
  \mathrm{CD} =
  \frac{1}{l}\!\left[
    \frac{1}{|\mathcal{B}^{\text{gen}}|}\!\!
      \sum_{p\in\mathcal{B}^{\text{gen}}}\!\!
        \min_{q\in\mathcal{B}^{\text{gt}}} \|p-q\|_2
    \;+\;
    \frac{1}{|\mathcal{B}^{\text{gt}}|}\!\!
      \sum_{q\in\mathcal{B}^{\text{gt}}}\!\!
        \min_{p\in\mathcal{B}^{\text{gen}}} \|p-q\|_2
  \right],
\]
where the per-pixel boundary distances are normalized by
the square image's side length ($l=448$ px) into a dimension-agnostic score for the ease of reproducibility.
This metric is the lower the better. 

\label{app:vbench}
\paragraph{$\Delta$VBench (Temporal Flickering, Motion Smoothness, Dynamic Degree).}
All three VBench~\citep{zhang2024vbench} metrics (temporal flickering, motion smoothness, dynamic degree) used in EgoX~\citep{egox2025} are \textbf{no-reference metrics
that scores a video without comparing to the ground-truth}, and real ego videos do not maximize these scores. The ground-truth videos actually score
worse than several of the baselines, which indicates flawed measurement.
Therefore, scoring higher on these metrics do not indicate better fidelity.
In this section, we report the metrics as delta deviation from GT, \textit{i.e.}
$|\text{score} - \text{GT score}|$. Comparing to VBench, T-LPIPS natively
compares against GT, thus we use it for the main table.

\begin{table}[h]
  \centering
  \footnotesize
  \setlength{\tabcolsep}{6pt}%
  \begin{tabular}{l c c c c c c}
    \toprule
    & \multicolumn{3}{c}{\textbf{New Action}}
    & \multicolumn{3}{c}{\textbf{Unseen}}\\
    \cmidrule(lr){2-4}\cmidrule(lr){5-7}
    Method
      & $\Delta_{\text{TF}}$$\downarrow$
      & $\Delta_{\text{MS}}$$\downarrow$
      & $\Delta_{\text{DD}}$$\downarrow$
      & $\Delta_{\text{TF}}$$\downarrow$
      & $\Delta_{\text{MS}}$$\downarrow$
      & $\Delta_{\text{DD}}$$\downarrow$\\
    \midrule
    Vista4D~\citep{lin2026vista4d} & $0.013$ & $0.004$ & $0.072$ & $0.010$ & $0.003$ & $0.160$\\
    EgoX~\citep{egox2025} (paper)  & $0.008$ & $0.002$ & $0.137$ & $0.013$ & $0.004$ & $0.209$\\
    \textbf{Ours}                  & $\textbf{0.004}$ & $\textbf{0.001}$ & $\textbf{0.003}$ & $\textbf{0.005}$ & $\textbf{0.001}$ & $\textbf{0.010}$\\
    \bottomrule
  \end{tabular}
  \caption{\textbf{$\Delta$VBench on the New-Activity (seen) and
    Unseen splits.} Lower magnitude is better; $0$ indicates a
    perfect match to GT on the corresponding signal.}
  \label{tab:vbench-supp}
\end{table}

\subsection{Variance across generations}
\label{app:seedvar}

Numbers in the main paper are averaged over five independent generations.
Table~\ref{tab:seedvar} reports the standard deviation across those runs.

\begin{table}[h]
  \centering
  \footnotesize
  \setlength{\tabcolsep}{4pt}%
  \resizebox{\textwidth}{!}{%
  \begin{tabular}{ll ccc ccc cc}
    \toprule
    & Variant & Loc Err & IoU & Contour & PSNR & SSIM & LPIPS & FVD & T-LPIPS \\
    \midrule
    \multirow{6}{*}{\rotatebox{90}{New Action}}
    & EgoX (LTX, data)$^{*}$        & $\pm0.001$ & $\pm0.003$ & $\pm0.001$ & $\pm0.03$ & $\pm0.001$ & $\pm0.002$ & $\pm7.6$ & $\pm0.023$ \\
    & EgoX (LTX, data, reloc)$^{*}$ & $\pm0.002$ & $\pm0.006$ & $\pm0.002$ & $\pm0.03$ & $\pm0.001$ & $\pm0.002$ & $\pm8.2$ & $\pm0.038$ \\
    & Ours (no obj tok, mask, synth)& $\pm0.001$ & $\pm0.005$ & $\pm0.001$ & $\pm0.03$ & $\pm0.001$ & $\pm0.001$ & $\pm4.3$ & $\pm0.022$ \\
    & Ours (no obj mask, synth)  & $\pm0.000$ & $\pm0.004$ & $\pm0.001$ & $\pm0.03$ & $\pm0.001$ & $\pm0.001$ & $\pm8.6$ & $\pm0.031$ \\
    & Ours (no synth)        & $\pm0.000$ & $\pm0.005$ & $\pm0.001$ & $\pm0.04$ & $\pm0.001$ & $\pm0.002$ & $\pm4.1$ & $\pm0.022$ \\
    & \textbf{Ours}           & $\pm0.001$ & $\pm0.005$ & $\pm0.001$ & $\pm0.05$ & $\pm0.001$ & $\pm0.001$ & $\pm2.0$ & $\pm0.016$ \\
    \midrule
    \multirow{6}{*}{\rotatebox{90}{Unseen}}
    & EgoX (LTX, data)$^{*}$        & $\pm0.001$ & $\pm0.018$ & $\pm0.002$ & $\pm0.05$ & $\pm0.003$ & $\pm0.001$ & $\pm21.4$ & $\pm0.056$ \\
    & EgoX (LTX, data, reloc)$^{*}$ & $\pm0.001$ & $\pm0.008$ & $\pm0.000$ & $\pm0.06$ & $\pm0.003$ & $\pm0.002$ & $\pm10.0$ & $\pm0.066$ \\
    & Ours (no obj tok, mask, synth)& $\pm0.004$ & $\pm0.007$ & $\pm0.005$ & $\pm0.05$ & $\pm0.004$ & $\pm0.003$ & $\pm5.6$ & $\pm0.067$ \\
    & Ours (no obj mask, synth)  & $\pm0.002$ & $\pm0.010$ & $\pm0.002$ & $\pm0.05$ & $\pm0.003$ & $\pm0.003$ & $\pm14.6$ & $\pm0.061$ \\
    & Ours (no synth)        & $\pm0.003$ & $\pm0.004$ & $\pm0.003$ & $\pm0.05$ & $\pm0.002$ & $\pm0.003$ & $\pm10.2$ & $\pm0.060$ \\
    & \textbf{Ours}           & $\pm0.002$ & $\pm0.008$ & $\pm0.003$ & $\pm0.05$ & $\pm0.002$ & $\pm0.002$ & $\pm15.4$ & $\pm0.027$ \\
    \bottomrule
  \end{tabular}}
  \caption{\textbf{Variance across generations.} Standard deviation over five
    independent generations. Means are reported in Tables~\ref{tab:main}
    and~\ref{tab:ablation}.}
  \label{tab:seedvar}
\end{table}

\section{Robustness to imperfect geometry, masks, and ego pose}
\label{app:robust}

Our conditioning is assembled from three estimated quantities: the monocular
depth behind the ego prior, the object segmentation masks that gate the
semantic tokens, and the target ego camera pose. This section measures how the generation quality degrades as each of them is corrupted.

Every corruption is applied at evaluation, so each row differs from its baseline by exactly one perturbation. 
Depth and pose are perturbed before rendering, so a single
knob degrades the whole geometric and semantic pathway together: rasterization, the
reprojected UV field, the 3D positional encoding, occupancy, and the mask reprojection that rides the same field. Perturbing them further downstream would corrupt the render while the gate still read clean geometry, which is not
a failure mode that can occur in practice. 
Depth error is injected either as spatially smooth
(low-frequency) error or as independent per-pixel noise of the same magnitude.

Mask perturbations are done after reprojection, so erosion and dilation radii are in gate pixels. 

The uncorrupted baselines are those of Table~\ref{tab:main}. 

\begin{table}[t]
  \centering
  \footnotesize
  \setlength{\tabcolsep}{4pt}%
  \resizebox{\textwidth}{!}{%
  \begin{tabular}{ll ccc ccc cc}
    \toprule
    & & \multicolumn{3}{c}{\textbf{Object Metrics}}
    & \multicolumn{3}{c}{\textbf{Image Metrics}}
    & \multicolumn{2}{c}{\textbf{Video Metrics}} \\
    \cmidrule(lr){3-5} \cmidrule(lr){6-8} \cmidrule(lr){9-10}
    & Depth error
    & Loc Err$\downarrow$ & IoU$\uparrow$ & Contour$\downarrow$
    & PSNR$\uparrow$ & SSIM$\uparrow$ & LPIPS$\downarrow$
    & FVD$\downarrow$ & T-LPIPS$\downarrow$ \\
    \midrule
    \multirow{9}{*}{\rotatebox{90}{\textbf{Ours}}}
    & none (deployed) & 0.0518 & 0.4121 & 0.0437 & 16.05 & 0.4600 & 0.4673 & 385.8 & 1.485 \\
    & 2.5\,cm smooth & 0.0555 & 0.3980 & 0.0480 & 15.96 & 0.4579 & 0.4752 & 400.2 & 1.530 \\
    & 2.5\,cm per-pixel & 0.0552 & 0.3911 & 0.0464 & 15.91 & 0.4571 & 0.4782 & 400.6 & 1.562 \\
    & 5\,cm smooth & 0.0581 & 0.3827 & 0.0502 & 15.83 & 0.4540 & 0.4848 & 410.1 & 1.580 \\
    & 5\,cm per-pixel & 0.0611 & 0.3627 & 0.0525 & 15.75 & 0.4504 & 0.4905 & 410.0 & 1.582 \\
    & 10\,cm smooth & 0.0630 & 0.3533 & 0.0534 & 15.62 & 0.4472 & 0.4978 & 429.6 & 1.654 \\
    & 10\,cm per-pixel & 0.0681 & 0.3151 & 0.0595 & 15.40 & 0.4376 & 0.5089 & 429.9 & 1.665 \\
    & 20\,cm smooth & 0.0721 & 0.3036 & 0.0626 & 15.32 & 0.4367 & 0.5133 & 427.5 & 1.822 \\
    & 20\,cm per-pixel & 0.0773 & 0.2696 & 0.0684 & 15.06 & 0.4242 & 0.5254 & 452.8 & 1.830 \\
    \midrule
    \multirow{9}{*}{\rotatebox{90}{\scriptsize\shortstack[c]{EgoX (LTX,\\data, reloc)$^{*}$}}}
    & none & 0.0702 & 0.3156 & 0.0631 & 15.92 & 0.4489 & 0.4973 & 467.4 & 2.063 \\
    & 2.5\,cm smooth & 0.0737 & 0.3083 & 0.0662 & 15.94 & 0.4515 & 0.5008 & 470.2 & 2.160 \\
    & 2.5\,cm per-pixel & 0.0735 & 0.3038 & 0.0666 & 15.93 & 0.4513 & 0.5026 & 466.2 & 2.193 \\
    & 5\,cm smooth & 0.0732 & 0.3015 & 0.0652 & 15.92 & 0.4513 & 0.5044 & 472.5 & 2.233 \\
    & 5\,cm per-pixel & 0.0776 & 0.2887 & 0.0689 & 15.88 & 0.4506 & 0.5072 & 473.4 & 2.279 \\
    & 10\,cm smooth & 0.0781 & 0.2868 & 0.0718 & 15.83 & 0.4493 & 0.5121 & 488.4 & 2.350 \\
    & 10\,cm per-pixel & 0.0788 & 0.2767 & 0.0719 & 15.76 & 0.4474 & 0.5153 & 484.1 & 2.433 \\
    & 20\,cm smooth & 0.0815 & 0.2633 & 0.0753 & 15.71 & 0.4467 & 0.5227 & 497.7 & 2.651 \\
    & 20\,cm per-pixel & 0.0856 & 0.2478 & 0.0801 & 15.59 & 0.4425 & 0.5302 & 513.9 & 2.719 \\
    \bottomrule
  \end{tabular}}
  \caption{\textbf{Robustness to imperfect geometry.} Depth is corrupted before
    rendering, at four magnitudes and in two spatial profiles. \emph{Ours}
    degrades faster in relative terms but stays ahead in absolute terms on every
    object and video metric at every level; \emph{EgoX} overtakes only on the two
    pixel-fidelity metrics, PSNR and SSIM.}
  \label{tab:robust-geom}
\end{table}

\paragraph{Imperfect geometry.}
In Table~\ref{tab:robust-geom} we show robustness analysis under different levels of geometric noise. We compare to the strongest EgoX variant, i.e. our implementation of EgoX (LTX, data, reloc).

\begin{table}[t]
  \centering
  \footnotesize
  \setlength{\tabcolsep}{4pt}%
  \resizebox{\textwidth}{!}{%
  \begin{tabular}{ll ccc ccc cc}
    \toprule
    & & \multicolumn{3}{c}{\textbf{Object Metrics}}
    & \multicolumn{3}{c}{\textbf{Image Metrics}}
    & \multicolumn{2}{c}{\textbf{Video Metrics}} \\
    \cmidrule(lr){3-5} \cmidrule(lr){6-8} \cmidrule(lr){9-10}
    & Pose error (rot. / trans.)
    & Loc Err$\downarrow$ & IoU$\uparrow$ & Contour$\downarrow$
    & PSNR$\uparrow$ & SSIM$\uparrow$ & LPIPS$\downarrow$
    & FVD$\downarrow$ & T-LPIPS$\downarrow$ \\
    \midrule
    \multirow{4}{*}{\rotatebox{90}{\textbf{Ours}}}
    & none (deployed) & 0.0518 & 0.4121 & 0.0437 & 16.05 & 0.4600 & 0.4673 & 385.8 & 1.485 \\
    & $1^\circ$ / $1\%$ ($0.017$\,m) & 0.0541 & 0.4009 & 0.0461 & 15.96 & 0.4556 & 0.4717 & 400.5 & 1.480 \\
    & $2^\circ$ / $2\%$ ($0.034$\,m) & 0.0538 & 0.3887 & 0.0461 & 15.85 & 0.4507 & 0.4774 & 398.5 & 1.493 \\
    & $5^\circ$ / $5\%$ ($0.085$\,m) & 0.0618 & 0.3254 & 0.0539 & 15.44 & 0.4360 & 0.5012 & 383.3 & 1.543 \\
    \midrule
    \multirow{4}{*}{\rotatebox{90}{\scriptsize\shortstack[c]{EgoX\\(LTX, data,\\reloc)$^{*}$}}}
    & none & 0.0702 & 0.3156 & 0.0631 & 15.92 & 0.4489 & 0.4973 & 467.4 & 2.063 \\
    & $1^\circ$ / $1\%$ ($0.017$\,m) & 0.0697 & 0.3185 & 0.0644 & 15.92 & 0.4501 & 0.4974 & 474.5 & 2.062 \\
    & $2^\circ$ / $2\%$ ($0.034$\,m) & 0.0724 & 0.3031 & 0.0650 & 15.85 & 0.4476 & 0.5010 & 473.5 & 2.073 \\
    & $5^\circ$ / $5\%$ ($0.085$\,m) & 0.0747 & 0.2742 & 0.0675 & 15.51 & 0.4385 & 0.5166 & 464.9 & 2.070 \\
    \bottomrule
  \end{tabular}}
  \caption{\textbf{Robustness to a noisy target ego pose.} One constant SE(3)
    offset per clip, applied before rendering: a mis-localized ego camera rather
    than per-frame jitter. Rotation is about a random axis. Translation is a
    fraction of that clip's median scene depth --- the reconstruction is
    scale-ambiguous, so a fixed offset in metres would be a different-sized
    error in every clip --- with the implied median offset in brackets (median
    scene depth is $1.69$\,m over \emph{Unseen}). Both methods are largely
    insensitive up to $2^\circ$ / $2\%$.}
  \label{tab:robust-pose}
\end{table}

\paragraph{Noisy ego pose.}
In Table~\ref{tab:robust-pose} we show robustness analysis under different levels of pose noise. We compare to the strongest EgoX variant, i.e. our implementation of EgoX (LTX, data, reloc).

\begin{table}[t]
  \centering
  \footnotesize
  \setlength{\tabcolsep}{4pt}%
  \resizebox{\textwidth}{!}{%
  \begin{tabular}{l ccc ccc cc}
    \toprule
    & \multicolumn{3}{c}{\textbf{Object Metrics}}
    & \multicolumn{3}{c}{\textbf{Image Metrics}}
    & \multicolumn{2}{c}{\textbf{Video Metrics}} \\
    \cmidrule(lr){2-4} \cmidrule(lr){5-7} \cmidrule(lr){8-9}
    Mask condition
    & Loc Err$\downarrow$ & IoU$\uparrow$ & Contour$\downarrow$
    & PSNR$\uparrow$ & SSIM$\uparrow$ & LPIPS$\downarrow$
    & FVD$\downarrow$ & T-LPIPS$\downarrow$ \\
    \midrule
    Ours & 0.0518 & 0.4121 & 0.0437 & 16.05 & 0.4600 & 0.4673 & 385.8 & 1.485 \\
    \midrule
    Given ego attn mask & 0.0461 & 0.4513 & 0.0381 & 16.30 & 0.4674 & 0.4529 & 373.6 & 1.462 \\
    \midrule
    erode 2\,px & 0.0528 & 0.4089 & 0.0452 & 16.01 & 0.4591 & 0.4708 & 401.3 & 1.492 \\
    erode 4\,px & 0.0557 & 0.3993 & 0.0479 & 15.99 & 0.4586 & 0.4714 & 397.7 & 1.499 \\
    erode 8\,px & 0.0543 & 0.4044 & 0.0467 & 15.99 & 0.4584 & 0.4715 & 397.0 & 1.499 \\
    dilate 2\,px & 0.0538 & 0.4050 & 0.0454 & 15.98 & 0.4568 & 0.4708 & 390.4 & 1.470 \\
    dilate 4\,px & 0.0571 & 0.3835 & 0.0486 & 15.90 & 0.4533 & 0.4758 & 392.2 & 1.454 \\
    dilate 8\,px & 0.0610 & 0.3687 & 0.0526 & 15.72 & 0.4469 & 0.4853 & 396.8 & 1.455 \\
    object tokens ungated & 0.0825 & 0.2945 & 0.0750 & 15.15 & 0.4224 & 0.5216 & 447.7 & 1.739 \\
    \bottomrule
  \end{tabular}}
  \caption{\textbf{Robustness to imperfect object masks} (\emph{Ours} only;
    EgoX has no object pathway). The bottom block varies how open the attention
    gate is, from tighter than deployed (erode), through looser (dilate), to
    removed altogether (ungated). Removing the gate costs more than any
    perturbation of it.}
  \label{tab:robust-mask}
\end{table}

\paragraph{Imperfect masks}
In Table~\ref{tab:robust-mask} we show robustness analysis under different levels of pose noise.

\section{Method details}
\label{app:method_details}

\subsection{Geometric Anchoring}
\label{app:geometric}

\paragraph{Mesh rasterization.}
Given re-localized ego poses and monocular depth, we build a 
triangle mesh from the exo depth grid by connecting neighboring pixels, cull
triangles at depth discontinuities (relative threshold 5\%), transform
vertices to the ego frame, and rasterize via
nvdiffrast~\citep{laine2020nvdiffrast}. This produces ego-view RGB
renderings, depth maps, and UV correspondence maps that the
subsequent stages consume.

\paragraph{Render-token injection.}
The geometric rendering tokens $\mathbf{Z}_{\mathrm{ego}}^{\mathrm{render}}$
(Eq.~\ref{eq:vae_encode}) are appended to the noisy target latent
$\mathbf{z}_{\sigma}$ rather than channel-concatenated. They share the same RoPE spatial coordinates as the noisy latent tokens, but their RoPE coordinates are x-shifted by the width of the ego video width.

\paragraph{Occupancy masking.}
\label{app:masking}
Rendered ego-view images contain holes (black regions) where the exo camera
has no coverage. After VAE encoding and patchification, some tokens cover
mostly-empty regions and carry no useful information. Furthermore, they could
mislead the model into generating black regions. Therefore, we generate token 
occupancy masks that filter out tokens that correspond to spatial temporal cells
with less than 10\% of the underlying pixels occupied.

\subsection{Semantic grounding}
\label{app:semantic}

\paragraph{Caption generation.}
\label{app:annotation}
We caption the first frame of each video split using a CapRL-Qwen3VL-4B~\citep{xing2025caprl}. Each caption contains
four sections: an overall scene description, up to 15 object noun phrases
describing visible objects, object spatial and interaction relations, and a predicted
ego-view description. The captioning covers all ego and exo cameras.

\paragraph{Object extraction.}
For each split we union the noun phrases across cameras with ego phrases prioritized. Gemma-3-12B~\citep{gemma2025gemma3} then returns up to 50 unique objects
ranked by relevance, with a single prompt per split, then deduplicates and re-ranks them.
The prompt instructs the model to merge phrases denoting the same object (e.g. "bike wheels" and "bicycle wheels" are considered the same), and to rank by relevance to the camera wearer's
viewpoint and interactions rather than by image area, so that objects the ego view
will actually contain are preferred over objects that merely occupy many exo
pixels. We use the top $K{=}10$ candidates. Because each phrase is used as a
text prompt for the segmentation model, which returns every instance matching that
phrase, the cap limits the number of distinct descriptions rather than the number
of segmented regions, for example, a single ``bike wheels'' slot may cover several wheels.

\paragraph{Obtaining ego segmentation masks from exo.}
The attention gates used by the cross-attention mechanism
(\S\ref{sec:semantic_grounding}) are per-object binary masks computed
from the exo view. We run SAM3~\citep{carion2025sam3} with the ranked
and deduplicated noun phrases as text prompts, producing per-object masks. 
Exo masks are first KB3-undistorted and then reprojected to the ego view.

\paragraph{Embedding object tokens.}
Each noun phrase $q_k$ is encoded by Gemma-3 as raw embeddings $G(q_k)$, padded to the fixed
per-object length $L{=}10$ tokens with learned register tokens $\mathbf{r}$, and
projected into the DiT cross-attention space by the trainable projector
$C_\phi$:
\begin{equation}
  \mathbf{O}_k
  =
  C_{\phi}\!\left(
    [\,G(q_k), \mathbf{r}_{|G(q_k)|},\ldots,\mathbf{r}_{L}\,]
  \right)
  \in \mathbb{R}^{L\times d}.
  \label{eq:object_tokens}
\end{equation}
The max length of the raw embeddings $G(q_k)$ across the dataset is 6, so
they are always padded.

\paragraph{Mixed-source mask training.}
During training we replace the reprojected exo masks with ground-truth ego segmentation masks with probability $p{=}0.5$.

\section{Data pipelines}
\label{app:data_pipelines}

\subsection{Real data pipeline}
\label{app:reconstruction}

\paragraph{The misalignment problem.}
EgoExo4D provides ground-truth ego camera poses and exo camera poses calibrated
with respect to the real world geometry. However, monocular depth estimators
produce 3D reconstructions whose scale and geometry do not match the real world. Rendering from the GT ego pose often places the scene at the wrong position. Such renderings provide unreliable signals that are usually misaligned or completely blank.

\paragraph{Re-localizing within the monocular reconstruction.}
Because the monocular reconstructions are inaccurate, we need to re-localize the ego
camera inside the monocular reconstruction to achieve good alignment between the
rendering and the ground-truth videos. 
Concretely, we build a per-clip alignment between MoGe-2~\citep{wang2025moge2}
reconstructions and a multi-view reconstruction from Depth-Anything-3~\citep{lin2025depthanything3}. 
Notice that the DA3 reconstruction is metric aligned because it uses ground-truth exo camera calibration, and it is only used to calculate the ego camera transformation, and we never use it during training or testing.
We then compute a non-rigid alignment from DA3's depth map to MoGe2.
The alignment is non-rigid in the sense that the depth alignment function is a depth-dependent alignment function. Please refer to the released code base for more details.
We then apply this alignment to the GT ego camera, bringing it from the real-world
coordinate frame into the MoGe2 reconstruction.
Only the depth of the camera is refined (and hence the 3D camera center), and the GT ego rotations are unchanged.
The result, used everywhere in training and evaluation, is a set of
ego poses that, when rendered from inside the monocular
reconstruction, produces ego-view images better aligned to the GT ego view.

Table~\ref{tab:reloc-shift} reports how far the camera relocalization step moves the ego
camera, measured over $14.0$M EgoExo4D frames. Table~\ref{tab:ablation} shows that the correction is worthwhile: adding it alone (\emph{EgoX (LTX, data, reloc)}$^{*}$, our re-implementation) improves all nine metrics.

\begin{table}[h]
  \centering
  \footnotesize
  \setlength{\tabcolsep}{10pt}%
  \begin{tabular}{l c c}
    \toprule
    Ego depth in exo view (m) & Shift (cm) & Shift (\%)\\
    \midrule
    $0$--$1$ & $4.5$   & $5.2$\\
    $1$--$3$ & $6.9$   & $3.9$\\
    $3$--$6$ & $83.0$  & $17.6$\\
    $>6$     & $185.5$ & $24.7$\\
    \bottomrule
  \end{tabular}
  \caption{\textbf{Ego-camera displacement from re-localization.}
    Median over $14.0$M frames, binned by the ego camera's depth along the
    exocentric optical axis. Shift (\%) is relative to the ego camera's
    distance from the exo camera.}
  \label{tab:reloc-shift}
\end{table}

\paragraph{Coordinate consistency, not depth repair.}
It is worth stating explicitly what the camera relocalization step does not attempt to correct the reconstruction and make it match the groundtruth video perfectly. This is because the reconstructions generated by off-the-shelf depth estimators are usually non-rigidly distorted, and no repositioning of the ego camera will
make the rendering exactly match the ground-truth ego frame everywhere. Our method does
not attempt this. The failure it removes is a coordinate-frame failure. The
EgoExo4D ego pose is calibrated against the real scene, whereas MoGe2 produces an
independently scaled and non-rigidly distorted proxy, so the calibrated camera is
not a valid camera in the proxy's frame at all; applying it directly places the
virtual camera in the wrong part of the reconstruction and yields a displaced or
nearly blank rendering regardless of how good the local geometry is.

An alternative would be to align the monocular depth itself to the multi-view
reconstruction and train on the corrected geometry. We deliberately do not do
this. Ground-truth depth is unavailable at inference time, creating a train/test mismatch. Training on the
original MoGe2 reconstruction instead exposes the generator to the same noise and
distortion characteristics it will see at test time, so that completion and
correction are learned jointly. The cost of this choice is that the learned
correction behavior is to some extent tuned to the error characteristics of
the reconstruction model, which we list as a limitation in
Appendix~\ref{app:limitations}.

\paragraph{Why this generalizes to in-the-wild data.}
Our model only ever observes renderings from the original MoGe2
reconstruction during training or testing. Therefore, as long as the ego camera can be localized within the MoGe2 reconstruction during test time, $e.g.$ via face tracking or manual view selection, our model can generate ego views in the same way
as trained.

\subsection{Generative synthetic data engine}
\label{app:syntheticdataset}

Our synthetic dataset uses randomized combinations of interior scenes, characters and animations. For each combination we render images and depth maps for ego-centric cameras, an exocentric camera, and save the camera world transform and field of view as meta data. The generation process works as follows:

We first create a set of interior scenes using Infinigen~\citep{raistrick2023infinigen} (640 in total).
We also create prompted, rigged, and textured characters through a customized character pipeline (550 in total), along with a wide variety of animations (3,000 in total, 30 fps) drawn from the Embody3D~\citep{embody3d} motion capture dataset, which records real human-human social interactions among groups of two to four people.

We then create random selections of pre-generated interior scenes, one or two characters, and assign a random set of character animations.
We then select a random 100-frame region in the animation timeline.
Ego cameras are then attached to the character's neck bone, therefore following the character animation. We add a temporal smoothing filter to reduce unnatural camera jitter, which is common to these captured sequences.
We then search for a valid exo camera through a randomized placement strategy.
We consider the exo camera to be in a valid configuration if at least 50\% of the joint positions of the characters are free of occlusions at all times in the selected animation range. At the same time the exo camera smoothly follows the average locations of the characters.
If no valid configuration could be found after 1000 iterations we discard the configuration.
After the exo camera has been placed in a valid location.
Using the described strategy, we sample 10,000 sequences from an asset pool.
We render the scenes through high-quality path tracing in Blender~\citep{blender} and write final color and metric depth passes. The ego cameras use fisheye projection while the exo camera uses perspective projection.

Please refer to our released code base for the actual implementation.

\section{Vista4D coordinate alignment}
\label{app:vista4d}

Vista4D~\citep{lin2026vista4d} is a conventional novel-view-synthesis
video model that targets
arbitrary 4D re-shooting from a casually captured monocular input.

\paragraph{Metric depth.}
Vista4D's reconstruction depth is up-to-scale, which is incompatible with metric
GT ego poses. Thus, similar to our own data pipeline, we calculate
a non-rigid transform that aligns its reconstruction with same DA-3 depth
used in our own pipeline (Appendix~\ref{app:reconstruction}). This alignment
is then used to locate the ego camera inside Vista4D's reconstruction.

\paragraph{Aria fisheye target.}
Vista4D is trained with pinhole cameras, but we find that it is 
generalizable to fisheye cameras. While the fisheye distortion is not
always accurate, the viewpoint and contents of the scenes are consistent
across fisheye and pinhole conditionings. We observe the same inability 
to generate ego views correctly.

\section{Limitations and societal impact}
\label{app:limitations}

\paragraph{Limitations.}
Several caveats apply to the deployed model.
(i)~\emph{Limited resolution and temporal length.}
The output egocentric video is generated at $448 \times 448$ over
$F = 33$ frames. Longer or higher-resolution synthesis would require chunk-by-chunk generation. \textbf{On our website, we show the latest chun-autoregressive results.}
(ii)~\emph{A target ego camera trajectory needs to be estimated at test time.}
The method synthesizes the ego view for a given ego camera path rather than inferring where the camera wearer will look. 
In our in-the-wild pipeline, this trajectory is produced automatically from human and face pose estimation
together with depth estimation, so the ego camera estimation accuracy bounds the accuracy of the view synthesis result.
(iii)~\emph{First-frame masks broadcast over time.}
The deployed model computes the object gate from segmentation of the first frame
and reuses that spatial gate for the remainder of the clip. For sequences with
substantial camera or object motion the gate becomes progressively stale, and
per-frame masks would be the natural remedy, though at a higher processing and storage cost before training can start.
(iv)~\emph{Dependence on semantic segmentation.}
We use off-the-shelf models for semantic segmentation. If the masks are too permissive, object semantics can be routed outside
the objects they describe and can potentially undermine generation. 
(v)~\emph{Dependence on 3D reconstruction.}
When the exocentric depth is severely incomplete, corrupted, or the ego view observes contents not shown in the exo video, video generation quality can be undermined.
(vi)~\emph{Dependence on large language models.}
The pipeline also leverages off-the-shelf VLM and LLM for captioning.
This makes the system capable but heavy, and reproducing it requires access to all of these components.
(vii)~\emph{Evaluation breadth.}
Following EgoX~\citep{egox2025} convention, our quantitative evaluation is performed on subsets of the EgoExo4D dataset because the full test set is too large for modern models to finish video generation within a reasonable compute and time budget. The in-the-wild results we show are qualitative and are not a second
quantitative benchmark.
(viii)~\emph{Ego-Exo consistency.} Due to occlusions and the generative nature of our model, the generated ego video is not perfectly consistent with the input exo video. This challenge might be alleviated by generating in the 3D space instead of 2D videos. We leave this for future work.

\paragraph{Societal impact.}
The negative societal impacts of this work are scenarios common to
generative models, such as potential misrepresentation of one's
first-person viewpoint and potential use for fraud. The positive
societal impacts are described as the motivation of this work
(Section~\ref{sec:intro}).

\section{Safeguards}
\label{app:safeguards}

Given the fraud and misrepresentation risks noted in
Appendix~\ref{app:limitations}, exo-to-ego generation systems should
ship with provenance safeguards. A standard option is invisible
digital watermarking: embedding signatures in the generated pixels
that are imperceptible to the human eye but recoverable by specialized
detector models, allowing downstream tools to flag synthesized
first-person video.

\end{document}